\documentclass{acmsmall}
\usepackage{url,graphicx}

\acmYear{2016}
\acmMonth{6}
\usepackage[ruled]{algorithm2e}
\SetAlFnt{\algofont}
\SetAlCapFnt{\algofont}
\SetAlCapNameFnt{\algofont}
\SetAlCapHSkip{0pt}
\IncMargin{-\parindent}

\title{A probabilistic tour of visual attention and gaze shift computational models}
\author{GIUSEPPE BOCCIGNONE  \affil{PHuSe Lab, Department of Computer Science, University of Milan, Milano, Italy} }

\begin{abstract}
In this paper  a number of problems are considered which are related to the  modelling of  eye guidance under visual attention in a natural setting. From a crude discussion of a variety of  available models spelled in probabilistic terms, it appears that current approaches in computational vision are hitherto far from achieving the goal of an active observer  relying upon eye guidance to accomplish real-world tasks. We argue that this challenging goal not only requires  to embody, in a principled way, the problem of eye guidance within the action/perception loop,  but to face the inextricable link tying up visual attention, emotion and executive control, in so far as recent neurobiological findings are weighed up. 
\end{abstract}

 
\begin{document}
 
 \begin{bottomstuff}
This  research  was  partially supported  by the project "Interpreting emotions: a computational tool integrating facial expressions and biosignals based shape analysis and bayesian networks",  grant FIRB - \emph{Future in Research} RBFR12VHR7\\
Author's address: G. Boccignone, Department of Computer Science, University of Milan, via Comelico 39/41, 20135 Milano, Italy; email: giuseppe.boccignone@unimi.it
\end{bottomstuff}
 
\maketitle

\section{Introduction}

As the french philosopher Merleau-Ponty put it, ``vision is a gaze at grips with a visible world'' \cite{maurice1945phenomenologie}. 
Goals and purposes, either internal or external, press the observer to maximise his information intake over time, by moment-to-moment sampling the most informative parts of the world.  In natural vision this endless endeavour  is accomplished through a sequence of  eye movements such as saccades and smooth pursuit, followed by fixations. Gaze shifts require visual attention to precede them to their goal, which has been shown to enhance the perception of selected part of the visual field (in turn related  to the foveal structure of the human eye, see \citeNP{Kowler2011} for an extensive discussion of these aspects).


The computational counterpart of using gaze shifts to enable a perceptual-motor analysis of the observed world can be traced back to pioneering work on active or animate vision \cite{aloimonos1988active,Ballard,bajcsy1992active}.  The main concern at the time was to embody vision in the action-perception loop of an artificial agent that purposively acts upon the environment, an idea that grounds its roots in early cybernetics \cite{cordeschi2002discovery}. To such aim the sensory apparatus  of the organism must be active and flexible, for instance, the vision system  can manipulate the viewpoint of the camera(s) in order to investigate the environment and get better information from it. 
Surprisingly enough, the link between  attention and active vision, notably when instantiated via gaze shifts (e.g, a moving camera),   was overlooked in those early approaches, as lucidly  remarked by \citeN{rothenstein2008attention}. Indeed, active vision, as it has been proposed and used in computer vision, must include attention as a sub-problem \cite{rothenstein2008attention}.  First and foremost when it must confront  the computational load  to achieve real-time processing (e.g., for autonomous robotics and videosurveillance).

Nevertheless, the mainstream of computer vision has  not dedicated to attentive processes and, more generally, to active perception  much consideration. This is probably due to the original sin of conceiving vision  as a pure information-processing task, a reconstruction process creating representations at increasingly  levels of abstraction, a land where action had no place: the ``from pixels to predicates'' paradigm \cite{aloimonos1988active}. 

To make a long story short, the research field had  a sudden burst when the paper by  \citeN{IttiKoch98}  was published. Their work provided a sound  and neat computational model (and the software simulation) to contend with the problem: in a nutshell, derive a saliency map  and generate gaze shifts as the result of  a Winner-Take-All (WTA) sequential selection of most salient locations. Since then, proposals and techniques have flourished. Under these circumstances, a deceptively simple question arises: Where are we now?

A straight answer, which is the \emph{leitmotiv} of this paper, is that whilst  early active vision approaches overlooked attention, current approaches have betrayed purposive active perception. 

In this perspective, here we provide a critical discussion of a number of models and techniques. It will be by no means exhaustive, and yet, to some extent, idiosyncratic. Our purpose  is not to offer a  review (there are  excellent ones the reader is urged to consult, e.g.,  \cite{BorItti2012,borji2014salient,bruce2015computational,bylinskii2015towards}), but rather to spell in a probabilistic framework the variety of  approaches, so to  discuss  in a principled way current limitations and to envisage intriguing directions of research, e.g., the hitherto neglected link between oculomotor behavior and emotion. 

In the following Section  we  highlight critical points  of current approaches and open issues. In Section \ref{sec:prob}  we frame such problems  in the language of probability. Section \ref{sec:action} discusses  possible routes to reconsider the problem of oculomotor behaviour within the action/perception loop.  In Section  \ref{sec:emo}  we explore the \emph{terra incognita} of gaze shifts and emotion.

\section{A Mini review and open issues}

The aim of a computational model of attentive eye guidance is to answer the question \emph{Where to Look Next?} by providing: 
\begin{enumerate}
\item at the \emph{computational theory level} (following  \citeNP{Marr}), an account of the mapping  from visual data of a  natural scene, say $\mathbf{I}$ (raw image data representing either a static picture or a stream of images), to  a sequence of  time-stamped gaze locations $(\mathbf{r}_{F_1}, t_1), (\mathbf{r}_{F_2}, t_2),\cdots$, 
namely 
\begin{equation}
\mathbf{I} \mapsto \{\mathbf{r}_{F_1}, t_1; \mathbf{r}_{F_2}, t_2;\cdots \}, 
\label{eq:mapping}
\end{equation}
\item at  the \emph{algorithmic level}, a procedure that simulates such mapping.
\end{enumerate}
 
\begin{figure}[h]
\centering
\includegraphics[scale=0.4,keepaspectratio=true]{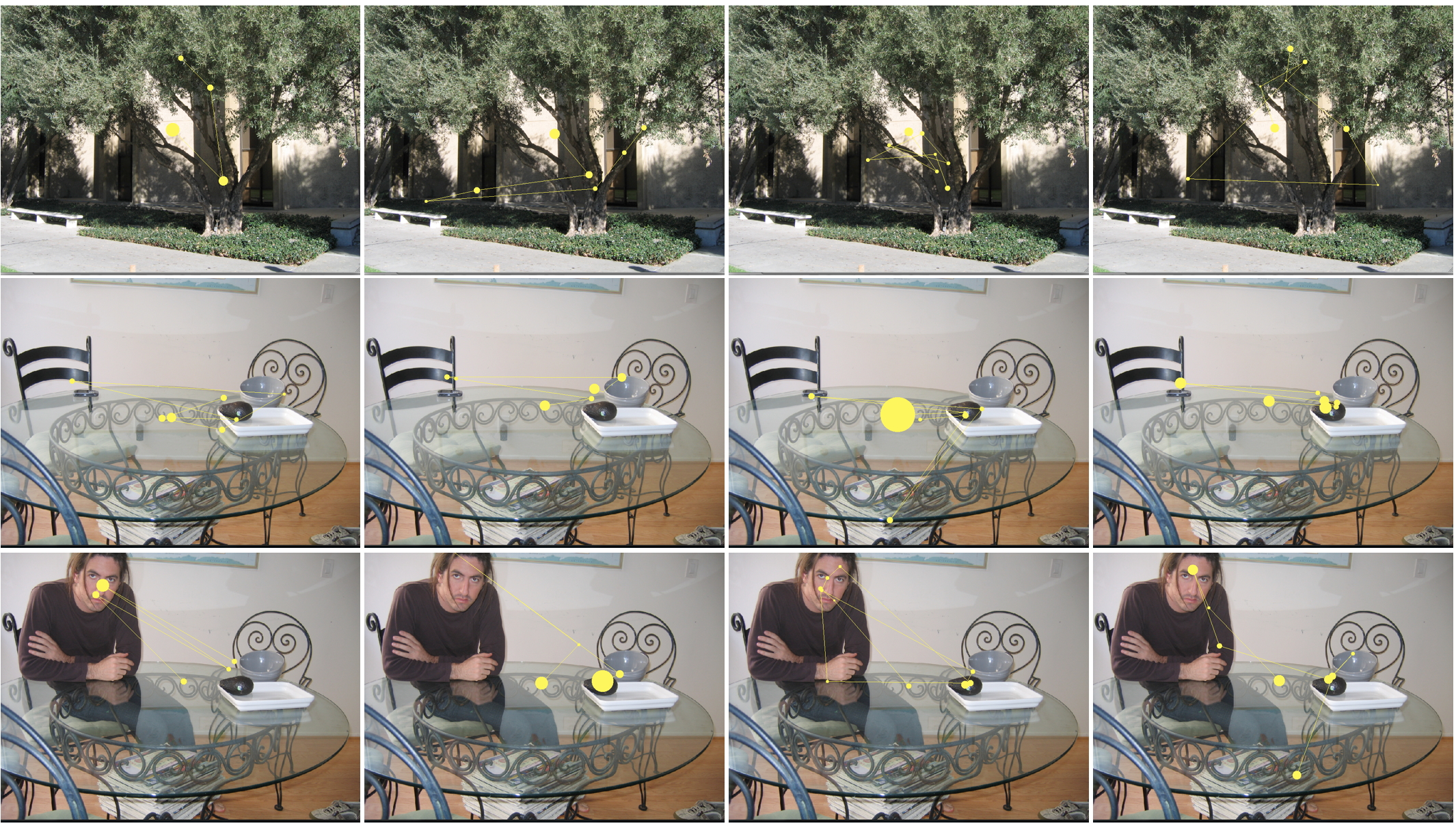}
\caption{Scan paths eye tracked from different human observers while viewing  three pictures of different information content: outdoor (top row), indoor with meaningful objects (middle row), indoor with high semantic content (person and face, bottom row). The area of yellow disks marking fixations between saccades is proportional to fixation time (images freely available from  the \textsf{Fixations in FAces} dataset).}
\label{Fig:variab}
\end{figure}

A simple example of the problem we are facing is shown in Figure \ref{Fig:variab}.

Under this conceptualization, when  the input $\mathbf{I}$ is  a static scene (a picture),  the  fixation duration time and saccade  (lengths and  directions) sequence are the only  observables  of the underlying guidance mechanism. When $\mathbf{I}$ stands for a time varying scene (e.g. a video), pursuit needs to be taken  into account, too. We will adopt the generic terms of gaze shifts for either pursuit, saccades and fixational movements.

In the following,   for notational simplicity, we will write  the time series $\{\mathbf{r}_{F_1}, t_1; \mathbf{r}_{F_2}, t_2;\cdots \}$ as the sequence  $\{\mathbf{r}_{F}(1), \mathbf{r}_{F}(2),\cdots \}$, unless the expanded form is needed.  Also, we will generically refer to such   sequence as  scan path, though this term has a historically precise meaning in the eye movement literature \cite{privitera2006scanpath}. 




The common practice of computational approaches to derive  the mapping (\ref{eq:mapping})  is to conceive it as a two step procedure: 
\begin{enumerate}
\item obtaining a suitable perceptual representation $\mathcal{W}$, i.e., $\mathbf{I} \mapsto  \mathcal{W}$; 
\item using $\mathcal{W}$ to generate the scan path, $\mathcal{W} \mapsto  \{\mathbf{r}_{F}(1), \mathbf{r}_{F}(2),\cdots \}$.
\end{enumerate}
It is important to remark that  each gaze position $\mathbf{r}_{F}(t)$ sets a new field of view for  perceiving the world,  thus  $\mathcal{W}=\{\mathcal{W}_{\mathbf{r}_{F}(1)}, \mathcal{W}_{\mathbf{r}_{F}(2)},\cdots \}$ should be a time-varying representation, even in the case of  a static image input. This feedback effect of the moving gaze is hardly considered at the modelling stage \cite{zelinsky2008theory,TatlerBallard2011eye}. 


By overviewing the field \cite{TatlerBallard2011eye,BorItti2012,bruce2015computational,bylinskii2015towards}, computational modelling has been mainly concerned with the first step: deriving a representation $\mathcal{W}$, typically in the form of a salience map. Yet, such step has recently evolved in a parallel research program, in which gaze shift prediction and simulation is not the focus, but  salient object detection (for an in-depth review of this ``second wave'' of saliency-centered methods, see \citeNP{borji2014salient}).

The second step, that is $\mathcal{W} \mapsto \{\mathbf{r}_{F}(1), \mathbf{r}_{F}(2),\cdots \}$, which actually brings in  the question of \emph{how} we look rather than \emph{where},  is seldom taken into account. Surprisingly, in spite of the fact that the  most cited work in the field  \cite{IttiKoch98} clearly addressed the \emph{how} issue  (gaze shifts as the result of  a  WTA sequential selection of most salient locations), most  models simply overlook the problem.
As a matter of fact, the  representation   $\mathcal{W}$, once computed, is usually validated in respect of its capacity for predicting the image regions that would be explored by  the overt attentional shifts of human observers (in a task designed to minimize the role of top-down factors). Predictability is assessed according to some established evaluation measures (see \citeNP{BorItti2012}, and  \citeNP{kummerer2015information}, for a recent discussion).  

In other cases,  if needed for practical purposes, e.g. for robotic applications, the  problem  of oculomotor action selection is solved by adopting some simple deterministic choice procedure  that  usually relies  on  selecting  the gaze position  $\mathbf{r}$ as the argument that  maximizes a measure on the given representation $\mathcal{W}$. 

We will attack on the problem related to gaze shift generation in Sections \ref{sec:bias} and \ref{sec:var}. In the following Section we first discuss representational problems.

%
%
%
%

\subsection{Levels of representation and control}
\label{sec:levels}
The guidance  of eye movements is likely to be influenced by a hierarchy of several interacting control loops, operating at different levels of processing.  Each processing step exploits the most suitable representation   of the viewed scene for its own level of abstraction. \citeN{schutz2011eye}, in a plausible portrayal,  have sorted out the following representational levels: 1) \emph{salience}, 2) \emph{objects}, 3) \emph{values}, and 4) \emph{plans}. 

Up to this date, the majority of  computational models have retained a central place for low-level visual conspicuity    \cite{TatlerBallard2011eye,BorItti2012,bruce2015computational}.
The perceptual representation of the world  $\mathcal{W}$  is usually epitomized in the form of a spatial saliency map, which is  mostly derived bottom-up (early salience)  following    \cite{IttiKoch98}.

The weakness of the bottom-up approach has been largely discussed (see, e.g. \citeNP{TatlerBallard2011eye,foulsham2008,EinhauserSpainPerona2008}).
Indeed, the effect of early salience  on attention is likely to be  a correlational effect rather than  an  causal one \cite{foulsham2008}, \cite{schutz2011eye}.  
Few examples are provided in Fig. \ref{Fig:itti}, where, as opposed to human scan paths (in free-viewing conditions), the scan path   generated by using a salience-based representation \cite{IttiKoch98}  does not  spot semantically important objects (faces),  the latter not being detected as regions of high contrast in colour, texture and luminance with respect to other regions of the picture. 

Under these circumstances,  early saliency can be modulated to  improve its fixation prediction. \citeN{Torralba} has  considered prior knowledge on the typical spatial location of the search target, as well as  contextual information (the gist of a scene, \citeNP{rensink2000dynamic}).  Further, object knowledge can be used to top-down tune early salience.  In particular, when  dealing with faces,  a face detection step  \cite{cerf2008predicting},  \cite{postma2011}, \cite{marat2013improving}  or a prior for Bayesian integration with low level features \cite{bocc08tcsvt}, can provide a reliable  cue to complement early conspicuity maps. Indeed, faces  drive attention in a direct fashion \cite{cerf2009faces} and the same holds for text regions \cite{cerf2008predicting,BocCOGN2014}.
It has been argued that salience has only an indirect effect on attention by acting through recognised objects: observers attend to interesting objects  and salience contributes little extra information to fixation prediction \cite{EinhauserSpainPerona2008}. 
 As a matter of fact, in the real world, most fixations are on task-relevant objects and this may or may not correlate with the  saliency of regions of the visual array  \cite{canosa2009real,rothkopfBallard2007}. 
 Notwithstanding,    object-based information has been scarcely taken into account in computational models \cite{TatlerBallard2011eye}. There are of course exceptions to this state of affairs, most notable ones  those provided by  \citeN{Rao2002},  \citeN{Sun2008}, the Bayesian models discussed by  \citeN{Poggio2010} and   \citeN{borji2012object}.

The representational problem is just the light side of the eye guidance problem. When actual eye tracking data are considered,  one has to confront with the dark side:   regardless of  the perceptual input, scan paths exhibit both systematic tendencies and notable inter- and intra-subject variability. As \citeN{canosa2009real} put it, where we choose to look next at any given moment in time is not completely deterministic,  but neither is it completely random. 

\subsection{Biases in oculomotor behaviour}
\label{sec:bias}

Systematic tendencies or ``biases'' in oculomotor behaviour can be thought of as regularities that are common across all instances of, and manipulations to, behavioural tasks  \cite{tatler2008systematic,tatler2009prominence}.  In that case case useful information about how the observers will move their eyes can be found. One remarkable example is 
the amplitude distribution of saccades and microsaccades that typically exhibit a positively skewed, long-tailed shape \cite{TatlerBallard2011eye,dorr2010variability,tatler2008systematic,tatler2009prominence}. Other paradigmatic examples of systematic tendencies in scene viewing are: initiating  saccades in the horizontal and vertical directions more frequently than in oblique directions;  small amplitude saccades tending to be followed by long amplitude ones and vice versa  \cite{tatler2008systematic,tatler2009prominence}.

%

Indeed,  biases affecting the manner in which we explore scenes with our eyes are well known in the psychological literature (see \citeNP{le2016introducing} for a thorough review), albeit underexploited in computational models.  
Such biases  may arise from a number of sources. \citeN{tatler2009prominence} have suggested the following:
biomechanical factors, saccade flight time and landing accuracy, uncertainty, distribution of objects of interest in the environment, task parameters.
%
%
%

Understanding biases in  eye guidance can provide powerful new insights into the decision about where to look in complex scenes. In a remarkable study, \citeN{tatler2009prominence}  provided striking evidence that a model based solely on these biases and therefore blind to current visual information can outperform  salience-based approaches.
Further, the  predictive performance of a salience-based model can be improved from 
$56\%$ to $80\%$ by including the probability of gaze shift  directions and amplitudes. 

Failing to account properly for such characteristics results in scan patterns that are fairly different from those generated by human observers (which can be easily noticed in the example provided in Fig. \ref{Fig:itti}) and eventually in  distributions of saccade amplitudes and orientations that do not match those estimated from human eye behaviour.

\subsection{Variability}
\label{sec:var}

\begin{figure}[h]
\centering
\includegraphics[scale=0.45,keepaspectratio=true]{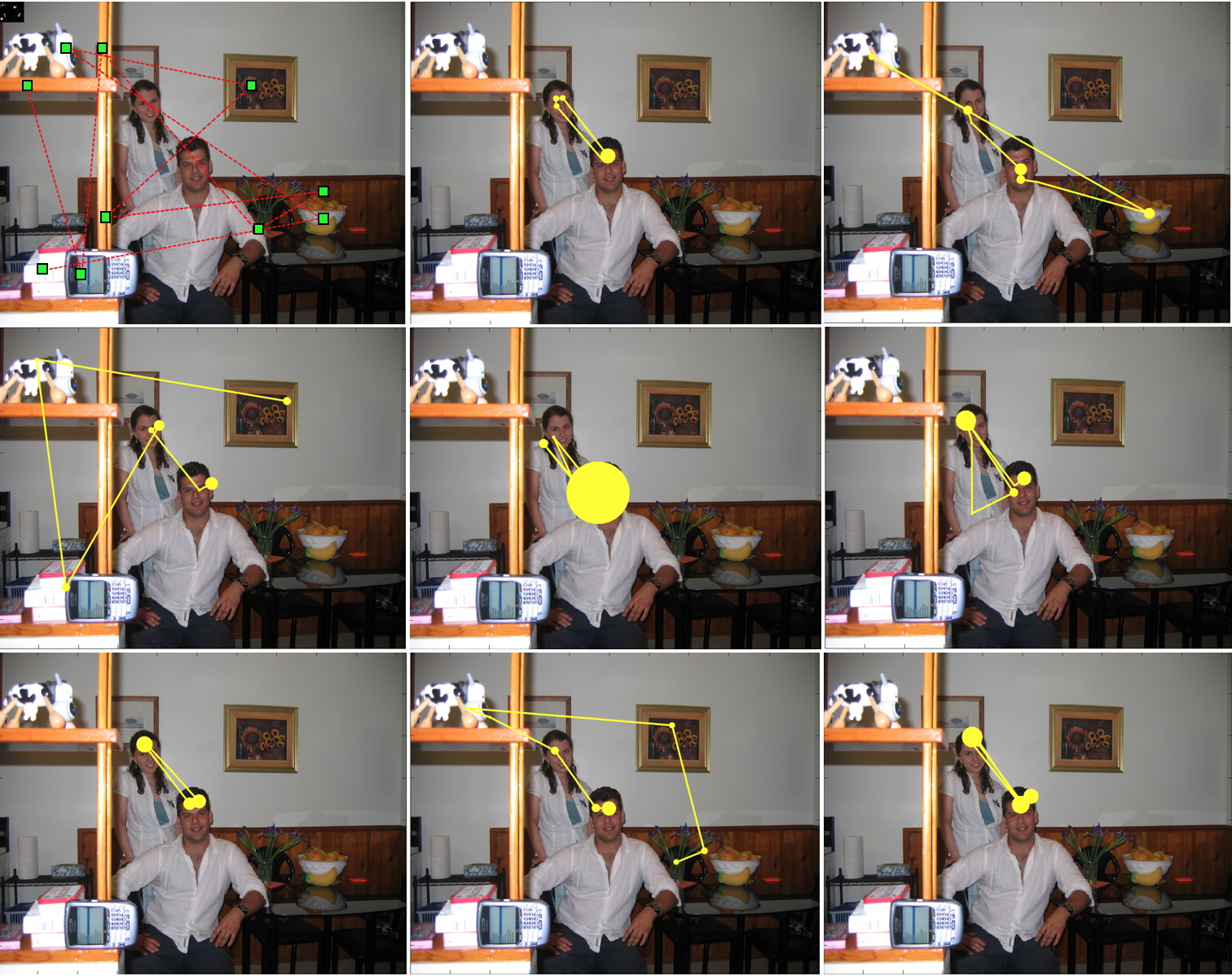}
\caption{A worst case analysis of scan path generation  based on early salience (top left image, via \citeNP{IttiKoch98})   vs.  scan paths eye tracked from human subjects. Despite of inter-subject variability concerning the gaze relocation pattern and fixation duration time, human observers consistently fixate on the two faces. The simulated scan path  fails to spot such semantically important objects that have not been highlighted as regions of high contrast in colour, texture and luminance. The area of yellow disks marking human fixations  is proportional to fixation time ( \textsf{Fixations in FAces} dataset).}
\label{Fig:itti}
\end{figure}

When looking  at natural images or movies \cite{dorr2010variability} under a free-viewing or a general-purpose task, the  relocation of gaze can be different among observers even though the same locations are taken into account.  In practice, there is a small probability  that two observers will fixate exactly the same location at exactly the same time. This effect is even more remarkable when free-viewing static images: consistency in fixation locations selected by observers decreases over the course of the first few fixations after stimulus onset \cite{TatlerBallard2011eye} and can become idiosyncratic.
 Such  variations in individual scan paths (as regards chosen fixations, spatial scanning order, and fixation duration)  still hold when the scene contains  semantically rich "objects" (e.g., faces, see Figures~\ref{Fig:variab} and \ref{Fig:itti}).  Variability  is  also exhibited by the same subject along different trials on equal stimuli.  

Randomness in motor responses is likely to be originated  from endogenous stochastic variations that affect each stage between a sensory event and the motor response: sensing, information processing, movement planning and executing \cite{vanBeers2007sources}.
It is worth noting that uncertainty comes into play since the earliest stage of visual processing: the human retina evolved such that high quality vision is restricted to the small part of the retina (about $2^{0}-5^{0}$ 
degrees of visual angle) aligned with the visual axis, the  fovea at the centre of vision. Thus, for many visually-guided behaviours the coarse information from peripheral vision is insufficient \cite{strasburger2011peripheral}. In certain circumstances, uncertainty may promote almost ``blind" visual exploration strategies \cite{tatler2009prominence,over2007coarse}, much like the behaviour of a foraging animal  exploring the environment under incomplete information; indeed when animals have limited information about where targets (e.g., resource patches) are located, different random search strategies may provide different chances to find them \citeNP{bartumeus2009optimal}.


Indeed, few works have been trying to cope with the variability issue, after the  early work by  \citeN{ellistark}, \citeN{hacisalihzade1992visual}. The  glorious WTA scheme  \cite{IttiKoch98}, or variants such as the selection of  the proto-object with the highest attentional weight  \cite{anna} are deterministic procedures.
Even when  probabilistic frameworks are used to infer  where to look next, the final decision is often taken via  the maximum a posteriori (MAP) criterion
which again is a deterministic procedure  (technically, an $\arg\max$ operation,  see \citeNP{elazary2010bayesian,bocc08tcsvt,geisler2005,ChernyakStark}), or  variants like the robust mean (arithmetic mean with maximum value) over candidate positions \cite{begum2010probabilistic}. As a result, for  a chosen visual  input $\mathbf{I}$
the mapping $\mathcal{W} \mapsto  \{\mathbf{r}_{F}(1), \mathbf{r}_{F}(2),\cdots \}$ will always generate the same scan path across different trials.

As a last remark, the variability of visual scan paths has been considered  a  nuisance rather than an opportunity from a modelling standpoint.  Nevertheless, beside  theoretical relevance for modelling human behavior, the randomness of the process can be an advantage in computer vision and learning tasks.  For instance,  \citeN{martinezLungarella} have reported  that a stochastic  attention selection mechanism (a refinement of the  algorithm  proposed in  \citeNP{bfpha04}) enables the i-Cub robot to explore its environment up to three times faster compared to the standard WTA  mechanism \cite{IttiKoch98}. Indeed, stochasticity makes the robot  sensitive to new signals and flexibly change its attention, which in turn enables efficient exploration of the environment as a basis for action learning \cite{nagai2009stability,nagai2009bottom}.

There are few notable exceptions to this current state of affairs, which will be discussed in Section \ref{sec:prior}.

\section{Framing models in a probabilistic setting}
\label{sec:prob}
We contend with the above issues by stating that observables such as fixation duration and  gaze shift lengths and directions are  random variables (RVs) that are generated by an underlying stochastic process. In other terms,  the sequence $\{\mathbf{r}_{F}(1), \mathbf{r}_{F}(2),\cdots \}$ is the realization of a stochastic process, and the ultimate goal of a computational theory  is to develop a mathematical model that describes statistical properties of  eye movements as closely as possible.
The problem of  answering the question \emph{Where to Look Next?} in a formal way  can be conveniently set in a probabilistic Bayesian framework. \citeN{tatler2009prominence} have re-phrased this question in terms of the posterior probability density function (pdf) $P(\mathbf{r} \mid \mathcal{W})$, which  accounts  for the plausibility of generating the gaze  shift $\mathbf{r} = \mathbf{r}_{F}(t) - \mathbf{r}_{F}(t-1)$, after the perceptual evaluation $\mathcal{W}$. Formally, via Bayes' rule:
\begin{equation}
P(\mathbf{r} \mid \mathcal{W})=  \frac{ P(\mathcal{W} \mid \mathbf{r} )}{P(\mathcal{W})}  P(\mathbf{r}).
\label{eq:BayesTatler}
\end{equation}

In Eq. \ref{eq:BayesTatler}, the first term on the r.h.s. accounts for the likelihood $P(\mathcal{W} \mid \mathbf{r})$ of $\mathbf{r}$ when visual data (e.g., features, such as edges or colors) are observed under a gaze shift  $\mathbf{r}_{F}(t) \rightarrow \mathbf{r}_{F}(t+1)$,    normalized by  $P(\mathcal{W})$, the evidence of the perceptual evaluation.  As they put it, ``The beauty of this approach is that the data could come from a variety of data sources such as simple feature cues, derivations such as Itti's definition of salience, object-or other high-level sources''.  The second term  is the pdf $P(\mathbf{r})$ incorporating prior knowledge on gaze shift execution. 

\begin{figure}[h!]
\label{Fig:tat}
\begin{minipage}[b]{.45\linewidth}
\centering\includegraphics[scale=0.5,keepaspectratio=true]{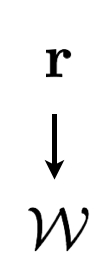}
\caption{ The generative model in PGM form supporting the Bayesian inference specified via Eq \ref{eq:BayesTatler}}
\label{fig:tata}
\end{minipage}%
\hspace{0.1\linewidth}
\begin{minipage}[b]{.45\linewidth}
\centering\includegraphics[scale=0.5,keepaspectratio=true]{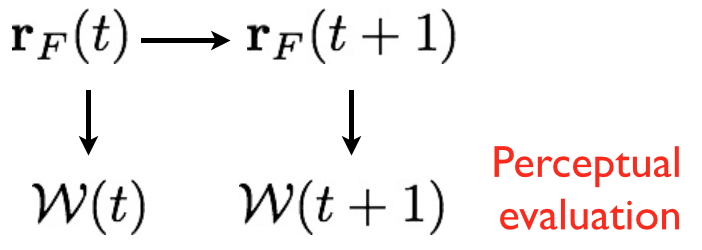}
\caption{The dynamic PGM obtained by unrolling in time the PGM depicted in Fig. \ref{fig:tata}}
\label{fig:tatb}
\end{minipage}
\end{figure}

The generative model behind  Eq. \ref{eq:BayesTatler}  is shown in Fig. \ref{fig:tata}   shaped in the form of a Probabilistic Graphical Model (PGM,  see \citeNP{murphy2012machine} for an introduction).  A PGM is a graph where nodes (e.g., $\mathbf{r}$ and  $\mathcal{W}$) denote RVs and  directed arcs (arrows)  encode conditional dependencies between RVs, e.g $P(\mathcal{W} \mid \mathbf{r})$. A node with no input arcs (for example $\mathbf{r}$) is associated with a prior probability, e.g., $P(\mathbf{r})$. Technically, as a whole, the PGM specifies at a glance a chosen factorization of the joint probability of all nodes. Thus, in Fig. \ref{fig:tata} we can promptly read that $P(\mathcal{W} , \mathbf{r}) = P(\mathcal{W} \mid \mathbf{r})P(\mathbf{r})$. The PGM in Fig. \ref{fig:tatb}  represents the PGM in Fig. \ref{fig:tata}, but unrolled in time. Note that now the arc $\mathbf{r}_{F}(t) \rightarrow \mathbf{r}_{F}(t+1)$ makes explicit the dynamics of the gaze shift occurring with probability $P(\mathbf{r}_{F}(t+1) \mid \mathbf{r}_{F}(t))$. 

The probabilistic model represented in Fig. \ref{Fig:tat} is generative in the sense that if all pdfs involved were fully specified, the attentive process could be simulated (via ancestral sampling, \citeNP{murphy2012machine}) as:
\begin{enumerate}
\item Sampling the gaze shift from the prior: 
\begin{equation}
\mathbf{r}^{*} \sim P(\mathbf{r});
\label{eq:priorTatlersamp}
\end{equation}
\item Sampling the observation of the world  under the gaze shift: 
\begin{equation}
\mathcal{W} ^{*} \sim P(\mathcal{W} \mid \mathbf{r}^{*}).
\label{eq:likeTatlersamp}
\end{equation}
\end{enumerate}

Inferring the gaze shift $\mathbf{r}$ when $\mathcal{W}$ is known boils down to the inverse probability problem (inverting the arrows), which is solved via Bayes' rule (Eq. \ref{eq:BayesTatler}). In the remainder of this paper we will largely use PGMs to simplify the presentation and discussion of probabilistic  models.

We will see in brief (Section \ref{sec:like}) that many current approaches previously mentioned can be accounted for by the likelihood term alone. But, crucial, and related to issues raised in  Section \ref{sec:bias}, is the Bayesian prior $P(\mathbf{r})$.

\subsection{The prior, first}
\label{sec:prior}

The  prior $P(\mathbf{r})$ can be defined \emph{prima facie} as the probability of shifting the gaze to a location \emph{irrespective of the visual information} at that location, although the term ``irrespective'' should be used with some caution \cite{le2016introducing}.  Indeed,  the prior is apt to encapsulate any systematic tendency in the manner in which we explore scenes with our eyes. The striking result obtained by  \citeNP{tatler2009prominence} is that if we  learn $P(\mathbf{r})$ from the actual observer's behavior, then we can stochastically sample gaze shifts (Eq.  \ref{eq:priorTatlersamp}) so to obtain  scan paths that, blind to visual information, out-perform feature-based accounts of eye guidance.



Note that the apparent simplicity of the prior term $P(\mathbf{r})$ hides a number of subtleties. For instance, Tatler and Vincent expand the  random vector  $\mathbf{r}$ in terms of its components, amplitude $l$ and direction $\theta$. Thus, $P(\mathbf{r})= P(l, \theta)$. This simple statement paves the way to different options. 

First easy option: such  RVs are marginally independent, thus,  $P(l, \theta) = P(l) P(\theta)$. In this case, gaze guidance, solely relying on biases, could be simulated by expanding Eq. \ref{eq:priorTatlersamp} via independent sampling of both components, i.e. at each time $t$, $l(t) \sim P(l(t)), \theta(t) \sim P(\theta(t))$. Alternative option: conjecture some kind of dependency, e.g. amplitude on direction so that $P(l, \theta) = P(l \mid \theta) P(\theta)$. In this case, the gaze shift sampling procedure would turn into the sequence $\widehat{\theta}(t) \sim P(\theta(t)), l(t) \sim P(l(t) \mid \widehat{\theta}(t) )$. Further:  assume that there is some persistence in the direction of the shift, which give rise to a stochastic process in which  subsequent directions are correlated, i.e., $\theta(t) \sim P(\theta(t) \mid \theta(t-1))$, and so on.

To summarize, by simply  taking into account the prior  $P(\mathbf{r})$, a richness of possible behaviors and analyses  are brought into the game. Unfortunately, most computational accounts of eye movements and visual attention have overlooked this opportunity,   with some exceptions. 
For instance, \citeNP{tavakoli2013stochastic} propose a system model for saccade generation in a stochastic filtering framework. A prior on amplitude $P(l(t))$ is considered by learning a Gaussian mixture model from eye tracking data. This way one aspect of biases is indirectly taken into account. It is not clear if their model accounts for variability and whether and how oculomotor statistics compare to human data.  In \cite{kimura2008dynamic},  simple eye-movements patterns are straightforwardly incorporated as a prior of a dynamic Bayesian network  to guide the sequence of eye focusing positions on videos.

In a different vein, \citeNP{le2016introducing}  have recently  addressed in-depth the bias  problem and made the interesting point that  viewing tendencies are not universal, but  modulated by the semantic visual category of the stimulus. They learn the joint pdf $P(l, \theta)$ of saccade amplitudes and orientations  via kernel density estimation; fixation duration is not taken into account. The model also brings in variability  \cite{le2015saccadic} by  generating a number  $N_c$ of random locations according to  conditional probability $P(\mathbf{r}_{F}(t) \mid \mathbf{r}_{F}(t-1))$ and the location with the highest saliency gain is chosen as the next fixation point. $N_c$ controls the degree of stochasticity.


Others have tried to capture eye movements randomness  \cite{keech2010eye1,rutishauser2007probabilistic}  but limiting to  specific tasks such as conjunctive visual search. A few more  exceptions can be found, but only  in the very peculiar field of eye-movements in reading  (see \citeNP{feng2006eye}, for a discussion). 

The variability and bias issues have been explicitly addressed from first principles in the theoretical context of L\'evy flights \cite{brockgeis,bfpha04}. The perceptual component was limited to a minimal core (e.g., based on a bottom-up salience map)  sufficient enough to support the eye guidance component. In particular in \cite{bfpha04} the long tail, positively skewed distribution of saccade amplitudes was shaped as a prior in the form of a Cauchy distribution,  whilst randomness was addressed at the algorithmic level  by prior sampling  $\mathbf{r} \sim P(\mathbf{r}(t))$  followed by  a Metropolis-like acceptance rule based on a deterministic saliency potential field.  The degree of stochasticity was controlled via the ``temperature'' parameter of the Metropolis algorithm.   The underlying eye guidance model was that of a random walker exploring the potential landscape (salience) according to a Langevin-like stochastic differential equation (SDE). The merit of such equation is the joint  treatment of both the deterministic  and the stochastic (variability) components behind eye guidance\footnote{Matlab simulation is available for download at \url{http://www.mathworks.com/matlabcentral/fileexchange/38512-visual-scanpaths-via-constrained-levy-exploration-of-a-saliency-landscape}}.

This basic mechanism has been refined and generalized in \cite{BocFerAnnals2012}  to composite $\alpha$-stable or L\'evy  random walks (the Cauchy law is but one instance of the class of $\alpha$-stable distributions), where, inspired by animal foraging behaviour,  a twofold regime can be distinguished: local exploitation (fixational movements following Brownian motion) and large exploration/relocation (saccade following L\'evy motion). What is interesting, with respect to the early model \cite{bfpha04}, is that the choice between the ``feed'' or ``fly'' states is made by sampling from a Bernoulli distribution, $Bern(z \mid \pi)$, with the parameter $\pi$   sampled from the conjugate  prior $Beta(\pi \mid \alpha, \beta)$. In turn, the behaviour of the Beta prior can be shaped via its hyperparameters $(\alpha, \beta)$, which, in an Empirical Bayes approximation,  can be tuned as a function of the class of perceptual data at hand (in the vein of \citeNP{le2016introducing}) and of time spent in feeding (fixation duration).
Most important, this approach paves the way  to the possibility of treating visual exploration strategies in terms  of \emph{foraging} strategies \cite{wolfe2013time,cain2012bayesian,BocFerSMCB2013,BocCOGN2014,napboc_TIP2015}. We will further expand on this in Section \ref{sec:action}.

\subsection{The unbearable lightness of  the likelihood}
\label{sec:like}


We noticed before, by inspecting Eq. \ref{eq:BayesTatler}  that the term  $\frac{ P(\mathcal{W} \mid \mathbf{r} )} {P(\mathcal{W})} $ could be related to many models proposed in the literature. This is an optimistic view. Most of the approaches actually discard the dynamics of gaze shifts implicitly captured by the shift vector $\mathbf{r}(t)$. In practice, they are more likely to be described  by a simplified version of Eq. \ref{eq:BayesTatler}: 
\begin{equation}
P(\mathbf{r}_{F} \mid \mathcal{W}) =  \frac{P(\mathcal{W} \mid \mathbf{r}_{F})} {P(\mathcal{W})}   P(\mathbf{r}_{F}).
\label{eq:BayesTatler2}
\end{equation}

The difference between Eq. \ref{eq:BayesTatler} and  \ref{eq:BayesTatler2} is subtle. The posterior $P(\mathbf{r}_{F} \mid \mathcal{W}) $ now answers the query ``What is the probability of \emph{fixating} at location $\mathbf{r}_{F}$ given visual data $\mathcal{W}$?''  Further, the prior  $P(\mathbf{r}_{F})$ simply accounts for the probability of spotting location $\mathbf{r}_{F}$.
 As a matter of fact, Eq. \ref{eq:BayesTatler2} bears no dynamics. 
 
\begin{figure}[h!]
\begin{minipage}[b]{.45\linewidth}
\centering\includegraphics[scale=0.3,keepaspectratio=true]{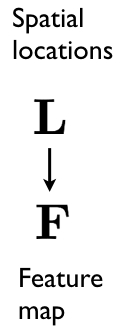}
\caption{PGM representing a feature-based  perceptual evaluation component $\mathcal{W}$}
\label{fig:simple}
\end{minipage}%
    \hspace{0.1\linewidth}
\begin{minipage}[b]{.45\linewidth}
\centering\includegraphics[scale=0.3,keepaspectratio=true]{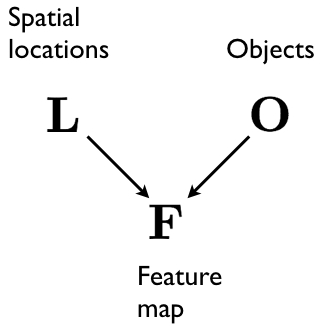}
\caption{PGM representing a object-based   perceptual evaluation component $\mathcal{W}$ \cite{torralba2006contextual}}
\label{fig:torralba}
\end{minipage}%
    \hspace{0.1\linewidth}
    \vspace{0.1\linewidth}
\begin{minipage}[b]{.45\linewidth}
\centering\includegraphics[scale=0.3,keepaspectratio=true]{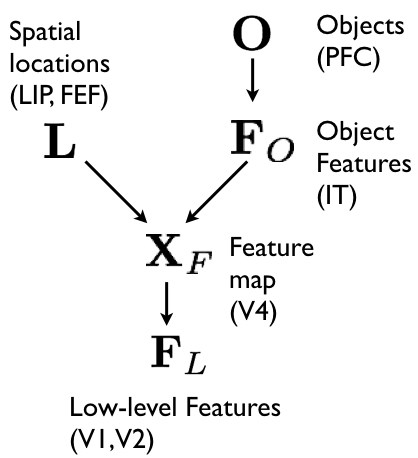}
\caption{The Bayesian model by  \citeN{Poggio2010},  which maps the PGM structure  to brain areas underpinning visual attention: early visual areas V1 and V2,  V4,  lateral intraparietal (LIP), frontal eye fields (FEF), inferotemporal (IT), prefrontal cortex (PFC)}
\label{fig:poggio}
\end{minipage}%
    \hspace{0.1\linewidth}
     \vspace{0.1\linewidth}
\begin{minipage}[b]{.45\linewidth}
\centering\includegraphics[scale=0.3,keepaspectratio=true]{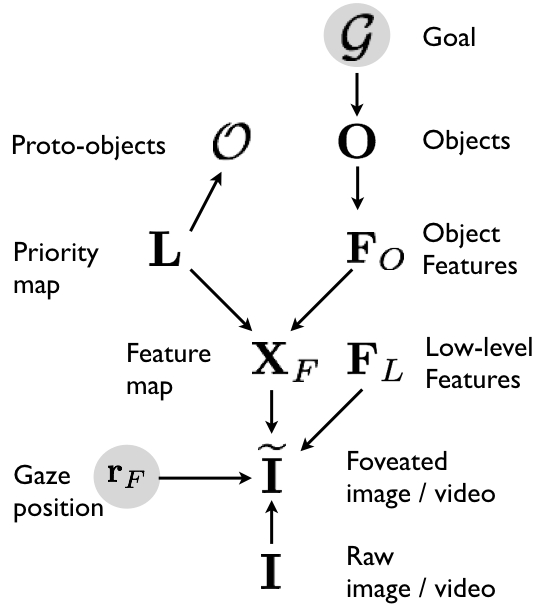}
\caption{An extension of  the PGM by \citeN{Poggio2010}, cfr. Fig. \ref{fig:poggio},  adopted in \cite{BocCOGN2014} and \cite{napboc_TIP2015}. In this  case $\mathcal{W}$ explicitly depends on the current gaze position $\mathbf{r}_{F}$  and goal $\mathcal{G}$.}
\label{fig:myPGM}
\end{minipage}
\label{Fig:models}  
\end{figure}

 In probabilistic terms we may re-phrase this result as the outcome of an assumption of independence: 
$P(\mathbf{r})   = P(\mathbf{r}_{F}(t) - \mathbf{r}_{F}(t-1)) \nonumber
  \simeq P(\mathbf{r}_{F}(t) \mid \mathbf{r}_{F}(t-1)) = P(\mathbf{r}_{F}(t))$.
To make things even clearer, let us explicitly substitute $\mathbf{r}_{F}$ with a RV $\mathbf{L}$ denoting locations in the scene, and $\mathcal{W}$ with RV $\mathbf{F}$ denoting features (whatever they may be);  then, Eq.~\ref{eq:BayesTatler2} boils down to 
\begin{equation}
P(\mathbf{L} \mid \mathbf{F}) =  \frac{P(\mathbf{F} \mid \mathbf{L})} {P(\mathbf{F})}  P(\mathbf{L}).
\label{eq:models}
\end{equation}
The PGM underlying  this inferential step is a very simple one and is represented in Figure~\ref{fig:simple}.  
A straightforward but principled use of Eq. \ref{eq:models}, which  has been exploited by approaches that draw upon   techniques  borrowed from statistical machine learning \cite{murphy2012machine} is the following:  consider $\mathbf{L}$ as a binary RV taking values in $\left[0,1\right]$ (or $\left[ -1, 1 \right]$), so that  $P(\mathbf{L} = 1 \mid \mathbf{F})$ represents the probability for a pixel, a superpixel or a patch  of being classified as  salient.

In the case  the  prior $P(\mathbf{L})$ is assumed to be uniform (no spatial bias, no preferred locations), then $ P(\mathbf{L} = 1 \mid \mathbf{F}) \simeq P(  \mathbf{F} \mid  \mathbf{L} = 1) $.  The likelihood function $P(  \mathbf{F} \mid  \mathbf{L} = 1)$ can be determined in many ways;  e.g.,  nonparametric kernel density estimation has been addressed  by \citeN{seo2009},  who use center / surround local regression kernels for computing  $\mathbf{F}$. 

More generally, taking into account the ratio 
$f(\mathbf{L})=\frac{P(\mathbf{L} = 1 \mid \mathbf{F})}{P(\mathbf{L} = 0 \mid \mathbf{F})}$ 
 (or, commonly, the log-ratio) casts the saliency detection problem in a classification problem, in particular a discriminative one \cite{murphy2012machine},   for which a variety of learning techniques are readily available. \citeN{kienzle2006nonparametric} pioneered this approach by learning the saliency discriminant function $f(\mathbf{L})$  directly from human eye tracking data using a support vector machine (SVM). Their approach has paved the way to a relevant number of works from \cite{judd2009learning} -- who trained a linear SVM from human fixation data using a set of low, middle and high-level  features to define salient locations--, to most recent ones that wholeheartedly  endorse machine learning trends.  Henceforth, methods have been proposed  relying on  sparse representation of ``feature words'' (atoms)   encoded in salient and non-salient dictionaries; these are  either learned from local image patches \cite{yan2010visual,lang2012saliency}  or  from eye tracking data of training images \cite{jiang2015image}. Graph-based learning is one other trend,  from the seminal work of \citeN{harel2007graph} to \citeN{yu2014maximal} (see the latter, for a brief review of this field).  Crucially, for the research practice, data-driven learning methods allow to contend with large scale  dynamic datasets.    \citeN{mathe2015actions} in the vein of \citeN{kienzle2006nonparametric} and \citeN{judd2009learning}  use SVM, but  they remarkably exploit state-of-the art computer vision datasets (Hollywood-2 and UCF Sports) annotated with human eye movements collected under the ecological constraints of a visual action recognition task.
 
As a general comment on (discriminative) machine learning-based methods, on the one hand it is embraceable the criticism  by \citeN{BorItti2012}, who surmise that these techniques make  ``models data-dependent, thus influencing fair model comparison, slow, and to some extent, black-box.''  But on the other hand, one important lesson of these approaches lies in that they  provides a data-driven way of deriving the most relevant visual features as optimal predictors. The learned  patterns  can shape receptive fields (filters) that have equivalent or superior predictive power when compared against hand-crafted (and sometimes more complicated) models \cite{kienzle2009center}. Certainly, this lesson is at the base of the current  exponentially growth of methods based on deep learning techniques \cite{lecun2015deep},  in particular Convolutional Neural Networks (CNN,  cfr. \citeNP{deepTaxonomy2016} for a focused review), where the computed features seem to outperform, at least from an engineering perspective, most of, if not all, the state-of-the art features conceived in computer vision. 

Again, CNNs,  as commonly exploited in the current practice, bring no significant conceptual novelty as to the use of Eq. \ref{eq:models}: fixation prediction is formulated  as a supervised  binary classification problem (in some case, regression  is addressed, \citeNP{wang2016deep}). For example,  \citeN{vig2014large}  use a linear SVM for learning the saliency discriminant function $f(\mathbf{L})$ after a large-scale search for optimal features $\mathbf{F}$. Similarly, \citeN{shenZhao2014learning} detect salient region  via linear SVM fed with features computed from multi-layer sparse network model. \cite{lin2014saliency} use the simple normalization step \cite{IttiKoch98} to  approximate $P(\mathbf{L} = 1 \mid \mathbf{F})$, where  \cite{kruthiventi2015deepfix} use the last  $1 \times 1$ convolutional layer of a fully convolutional net. Cogent here  is the outstanding performance of CNN in learning and representing  features that correlate well with eye fixations, like objects, faces, context. 
 
Clearly, one problem is  the enormous amount of training data necessary to train these networks, and the  engineering expertise required, which makes them difficult to apply  for predicting saliency.  However, \citeN{kummerer2014deep}  by exploiting the well known  network from \cite{AlexNetNIPS2012} as starting point,  have given evidence that deep CNN  trained on computer vision tasks like object detection boost saliency prediction. The network by \citeN{AlexNetNIPS2012} has been optimized for object recognition using a massive dataset consisting of more than one million images, and  results reported by \citeN{kummerer2014deep}  on static pictures are impressive when compared to state-of-the-art methods, even  to previous CNN-based proposals \cite{vig2014large}.

Apart from the straightforward implementation via popular  machine-learning algorithms,  the ``light'' model described by Eq. \ref{eq:models} is further amenable  to a minimal  model, which, surprisingly enough, is however capable of accounting for a large number of approaches. This can be easily appreciated by setting $P(\mathbf{F} \mid \mathbf{L}) = const.,  P(\mathbf{L})=const.$  so that Eq. \ref{eq:models} reduces to
\begin{equation}
P(\mathbf{L} \mid \mathbf{F}) \propto  \frac{1}{P(\mathbf{F})}.
\label{eq:modelItti}
\end{equation}  
Eq.~\ref{eq:modelItti} states that the probability of fixating a spatial location $\mathbf{L}= (x,y)$ is higher when ``unlikely'' features (unlikeliness $ \approx \frac{1}  {P(\mathbf{F})}$)  occur at that location. In a natural scene, it is typically the case of high contrast regions (with respect to either luminance, color, texture or motion). 
This is nothing but the   salience-based  component of the most prominent model in the literature \cite{IttiKoch98}, which Eq.~\ref{eq:modelItti} re-phrases in probabilistic terms.

A thorough reading of the  review by  \citeN{BorItti2012} is sufficient to gain the understanding that a great deal of computational models so far proposed (47 over 63 models)   are much or less variations  of this theme (albeit experimenting with different features, different weights for combining them, etc.) even when  sophisticated probabilistic techniques are adopted  to shape the distribution $P(\mathbf{F})$ (e.g., nonparametric Bayes techniques, \citeNP{BoccICPR08}). 
 Clearly, there are works that have tried to avoid weaknesses related to such a light-modelling of the perceptual input, and have tried to climb up the levels of the representation hierarchy \cite{schutz2011eye}. Some examples  are summarized at a glance in Figure~\ref{Fig:models} (but see \citeNP{BorItti2012}).
 
Nevertheless, in spite of its simplicity, Eq.~\ref{eq:modelItti} is apt to pave the way to interesting  frameworks. For instance, by noting that  $\log \frac{1}  {P(\mathbf{F})}$ is nothing but  Shannon's Self- Information, information theoretic approaches become available at the algorithmic level. These approaches  set computational constraints  under the general assumption that saliency computation serves to maximize information sampled from the environment \cite{bruce2009saliency}. 
 
Keeping on with  the information theory framework, and going back to Eq.~\ref{eq:models}, a simple manipulation,
\begin{equation}
\log P(\mathbf{L} \mid \mathbf{F}) - \log P(\mathbf{L}) =  \log \frac{P(\mathbf{F} \mid \mathbf{L})} {P(\mathbf{F})}  
\label{eq:modelsKL}
\end{equation}
\noindent sets the focus on the discrepancy, or dissimilarity, $\log P(\mathbf{L} \mid \mathbf{F}) - \log P(\mathbf{L}) = \log \frac{P(\mathbf{L} \mid \mathbf{F})}{P(\mathbf{L})}$ between the log-posterior and the log-prior. A (non-commutative) measure, formalizing this notion of dissimilarity is readily available in information theory, namely the  Kullback-Leibler (K-L) divergence between two distributions $P(X)$ and $Q(X)$ \cite{Mackay}:
\begin{equation}
D_{KL}(P(X) || Q(X))= \int_X  \log \frac{P(x)} {Q(x)} P(x) dx  
\label{eq:KL}
\end{equation}
Measuring differences between posterior and prior beliefs of the observers is however a general concept applicable across different levels of abstraction. For instance, one might consider the object-based model  \cite{torralba2006contextual}  in Fig. \ref{fig:torralba}, 
which can be used for inferring the joint posterior of gazing at  certain kinds of objects $\mathbf{O}$ at location $\mathbf{L}$ of a viewed scene, namely,    $P(\mathbf{O},\mathbf{L} \mid \mathbf{F}) \propto P(\mathbf{F} \mid \mathbf{O},\mathbf{L}) P(\mathbf{O},\mathbf{L})$.  Then, $D_{KL}(P(\mathbf{O},\mathbf{L} \mid \mathbf{F}) || P(\mathbf{O},\mathbf{L}))$ is the average of the log-odd ratio, measuring  the divergence between observer's prior belief distribution on  $(\mathbf{O},\mathbf{L})$ and his posterior belief distributions after perceptual data $\mathbf{F}$ have been gathered. Indeed, this is a statement that can be generalized to any model $\mathbf{M}$ in  a model space $\mathcal{M}$ and new data observation $\mathbf{D}$ so to define the Bayesian surprise \cite{balditti2010} $\mathcal{S}(\mathbf{D},\mathbf{M})$: $\mathbf{D}$  is surprising if the posterior distribution resulting from observing $\mathbf{D}$  significantly differs from the prior distribution, i.e., $S(\mathbf{D},\mathbf{M}) = D_{KL}(P(\mathbf{M} \mid \mathbf{D} ) || P(\mathbf{M}))$. \citeN{itti2009bayesian} have shown that Bayesian surprise attracts human attention in dynamic natural scenes. To recap, Bayesian surprise is a measure of salience based on the K--L divergence.

Eventually, note that the K-L divergence (\ref{eq:KL}) is a flexible tool and can be used for different purposes. For instance, when dealing with  models of perceptual evaluation such as those specified in Figs \ref{fig:torralba}, \ref{fig:poggio},  and
\ref{fig:myPGM}, once the model has been detailed at the computational theory level via its PGM, then using the latter for learning  inference and prediction brings  in the algorithmic level. Indeed, for any Bayesian generative model other than trivial ones, 
such steps are usually performed in approximate form \cite{murphy2012machine}. Stochastic approximation resorting to algorithms  such as   Markov-chain Monte Carlo (MCMC) and Particle Filtering (PF)  is one possible choice; the alternative choice is represented by deterministic optimization algorithms \cite{murphy2012machine} such as variational Bayes (VB) or belief propagation (BP, a message passing scheme exchanging beliefs between PGM nodes). For example, the model by  \citeN{Poggio2010},  following the work of \citeN{rao2005}, relies upon BP message passing for inferential steps. Interestingly enough, \citeN{rao2005} has argued for a plausible neural implementation of BP.   VB algorithms, on the other hand, are based on Eq. \ref{eq:KL}, where  $P$ usually stands for a complete distribution and $Q$ is the approximating distribution; then,  parameter (or model) learning is accomplished by minimizing  the K-L divergence  (as an example, the well known Expectation-Maximization algorithm, EM, can be considered a specific case of the VB  algorithm, \citeNP{Mackay,murphy2012machine}). In Section \ref{sec:action} we will also touch on a deeper interpretation of the  K-L  minimization / VB algorithm.

But at this point a simple question arises: where have the eye movements gone?

\section{Making a step forward: back to the beginning of active vision}
\label{sec:action}

Visual perception coupled with gaze shifts should be considered the  \emph{Drosophila} of  perception-action loops. Among the variety of active behaviors the organism can fluently engage to purposively act upon and perceive  the world (e.g, moving the body, turning the head, manipulating objects), oculomotor behavior is the minimal, least energy, unit. To perform $3$-$4$ saccades per second, the organism roughly spends $300$ msecs to close the loop  ($200$ msecs for motor preparation and execution, $100$ msecs left for perception).

\begin{figure}[t]
\centering
\includegraphics[scale=0.5,keepaspectratio=true]{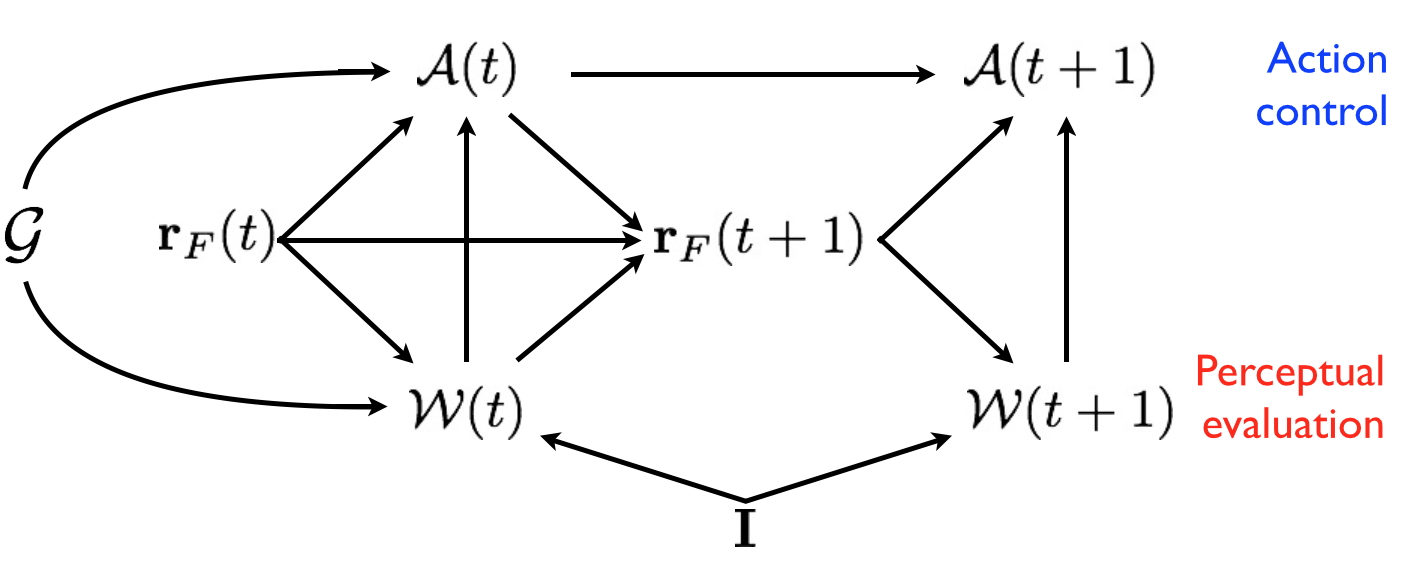}
\caption{The perception-action loop unfolded in time as a dynamic PGM. $\mathcal{A}(t)$ denotes the ensemble of time-varying RVs defining the oculomotor action setting; $\mathcal{W}(t)$  stands for the ensemble of time-varying RVs characterising  the scene as actively perceived  by the observer; $\mathcal{G}$ summarizes the given goal(s), To simplify the graphics, conditional dependencies $\mathcal{G} \rightarrow \mathcal{A}(t+1)$ and $\mathcal{G} \rightarrow \mathcal{W}(t+1)$ have been omitted.}
\label{Fig:loop}
\end{figure}

One way to make justice of this forgotten link is going back to first principles by re-shaping  the  problem as  the action-perception loop, which is presented in Figure \ref{Fig:loop} in the form of a dynamic PGM. The model relies upon the following assumptions:
\begin{itemize}
\item The  scene  that will be perceived at time $t+1$, namely $\mathcal{W}(t+1)$  is inferred  from the raw data  $\mathbf{I}$, gazed at  $\mathbf{r}_{F}(t+1)$, under the  goal  $\mathcal{G}$  assigned to the observer, and is conditionally dependent on current perception $\mathcal{W}(t)$. Thus, the perceptual inference problem is summarised by the conditional distribution  $P( \mathcal{W}(t+1)|\mathcal{W}(t), \mathbf{r}_{F}(t+1), \mathbf{I},\mathcal{G})$;
\item   The external goal  $\mathcal{G}$ being assigned,  the oculomotor action setting at time $t+1$,  $\mathcal{A}(t+1)$,  is drawn conditionally on current action setting $\mathcal{A}(t)$  and the  perceived scene  $\mathcal{W}(t+1)$ under gaze position $\mathbf{r}_{F}(t+1)$; thus,  its  evolution in time   is inferred according to the conditional distribution  $P(\mathcal{A}(t+1) | \mathcal{A}(t), \mathcal{W}(t+1), \mathbf{r}_{F}(t+1), \mathcal{G})$.
\end{itemize}
Note that the action setting dynamics $\mathcal{A}(t) \rightarrow \mathcal{A}(t+1)$ and the scene perception dynamics $\mathcal{W}(t) \rightarrow \mathcal{W}(t+1)$ are intertwined with one another by means of the gaze shift process $\mathbf{r}_{F}(t) \rightarrow \mathbf{r}_{F}(t+1)$: on the one hand  next gaze position $\mathbf{r}_{F}(t+1)$ is used to define a distribution on $\mathcal{W}(t+1)$ and $\mathcal{A}(t+1)$; meanwhile,  the probability distribution of  $\mathbf{r}_{F}(t+1)$ is conditioned on current gaze position,  $\mathcal{W}(t)$ and $\mathcal{A}(t)$, namely $P(\mathbf{r}_{F}(t+1)| \mathcal{A}(t), \mathcal{W}(t), \mathbf{r}_{F}(t))$.

We have previously  discussed the perceptual evaluation component $\mathcal{W}(t)$.  A general way of defining the oculomotor executive control component $\mathcal{A}(t)$ is through the following ensemble of RVs:
\begin{itemize}
\item $\{\mathbf{V}(t), \mathbf{R}(t)\}$:  $\mathbf{V}(t)$ is a spatially defined RV   used to provide a suitable  probabilistic representation of value;  $\mathbf{R}(t)$ is a binary RV defining  whether or not a payoff (either positive or negative) is returned;
\item $\{\pi(t), z(t),   \xi(t) \}$: an \emph{oculomotor state representation} as defined via the multinomial RV  $z(t)$, occurring with probability $\pi(t)$, and determining the choice of motor parameters $\xi(z,t)$ guiding the actual gaze relocation (e.g., lenght and direction of a saccade as opposed to those driving a smooth pursuit) ;
\item $\mathcal{D}(t)$:  a set of state-dependent statistical decision rules  to be applied on a set of candidate new gaze locations $\mathbf{r}_{new}(t+1)$ distributed according to the posterior pdf of $\mathbf{r}_{F}(t+1)$.
\end{itemize}

In the end,   the actual shift  can be summarised as the statistical decision of selecting   a  particular gaze location  $\mathbf{r}^{\star}_{F}(t+1)$   on the basis of $P(\mathbf{r}_{F}(t+1)| \mathcal{A}(t), \mathcal{W}(t), \mathbf{r}_{F}(t))$ so to maximize the expected payoff under the current goal $\mathcal{G}$, and  the  action/perception cycle   
boils down to the iteration of the following steps:
\begin{enumerate}
\item Sampling the gaze-dependent current perception:
\begin{equation}
\mathcal{W}^{*}(t) \sim P( \mathcal{W}(t)|\mathbf{r}_{F}(t),\mathbf{F}(t), \mathbf{I}(t),\mathcal{G});
\label{eq:step1}
\end{equation}
\item Sampling the appropriate motor behavior (e.g., fixation or saccade):
\begin{equation}
\mathcal{A}(t)^{*} \sim P(\mathcal{A}(t) | \mathcal{A}(t-1), \mathcal{W}^{*}(t),\mathcal{G});
\label{eq:step2}
\end{equation}
\item Sampling where to look next:
\begin{equation}
\mathbf{r}_{F}(t+1) \sim P(\mathbf{r}_{F}(t+1)| \mathcal{A}(t)^{*} , \mathcal{W}^{*}(t), \mathbf{r}_{F}(t)).
\label{eq:step3}
\end{equation}
\end{enumerate}

It is worth noticing that  we have chosen to describe the observer's action / perception cycle in terms of stochastic sampling based on the probabilistic  model  in Figure \ref{Fig:loop}. However,  one can recast the inferential problems in terms of deterministic optimization: in brief, optimising the probabilistic model $\mathcal{W}$ of how sensations are caused, so that the resulting predictions can select the optimal $\mathcal{A}$ to guide active sampling (gaze shift) of sensory data. One such approach, which is well known in theoretical neuroscience but, surprisingly,  hitherto unconsidered in computer vision, relies on the free-energy principle \cite{friston2010free,feldman2010attention,friston2013anatomy}.  Free-energy $\mathcal{F}$ is a quantity from statistical physics and  information theory \cite{Mackay} that bounds the negative log-evidence of sensory data.  Under simplifying assumptions, it boils down to the amount of prediction error of sensory data under  a model.  In such context, the action / perception cycle is the result of a dual minimization process: i) action  reduces  $\mathcal{F}$  by changing sensory input, namely by sampling  (via gaze shifts) what one expects consistent with perceptual inferences; ii) perception reduces $\mathcal{F}$ by making inferences about the causes of sampled sensory signals and changing predictions. Friston defines this process ``active inference''.  To make a connection with Section \ref{sec:like}, by minimizing the free-energy, Bayesian surprise is maximised; indeed, in Bayesian  learning, free energy minimization   is a common rationale behind many optimisation techniques such as VB and BP  \cite{Mackay,murphy2012machine}

\begin{figure}[t]
\centering
\includegraphics[scale=0.5,keepaspectratio=true]{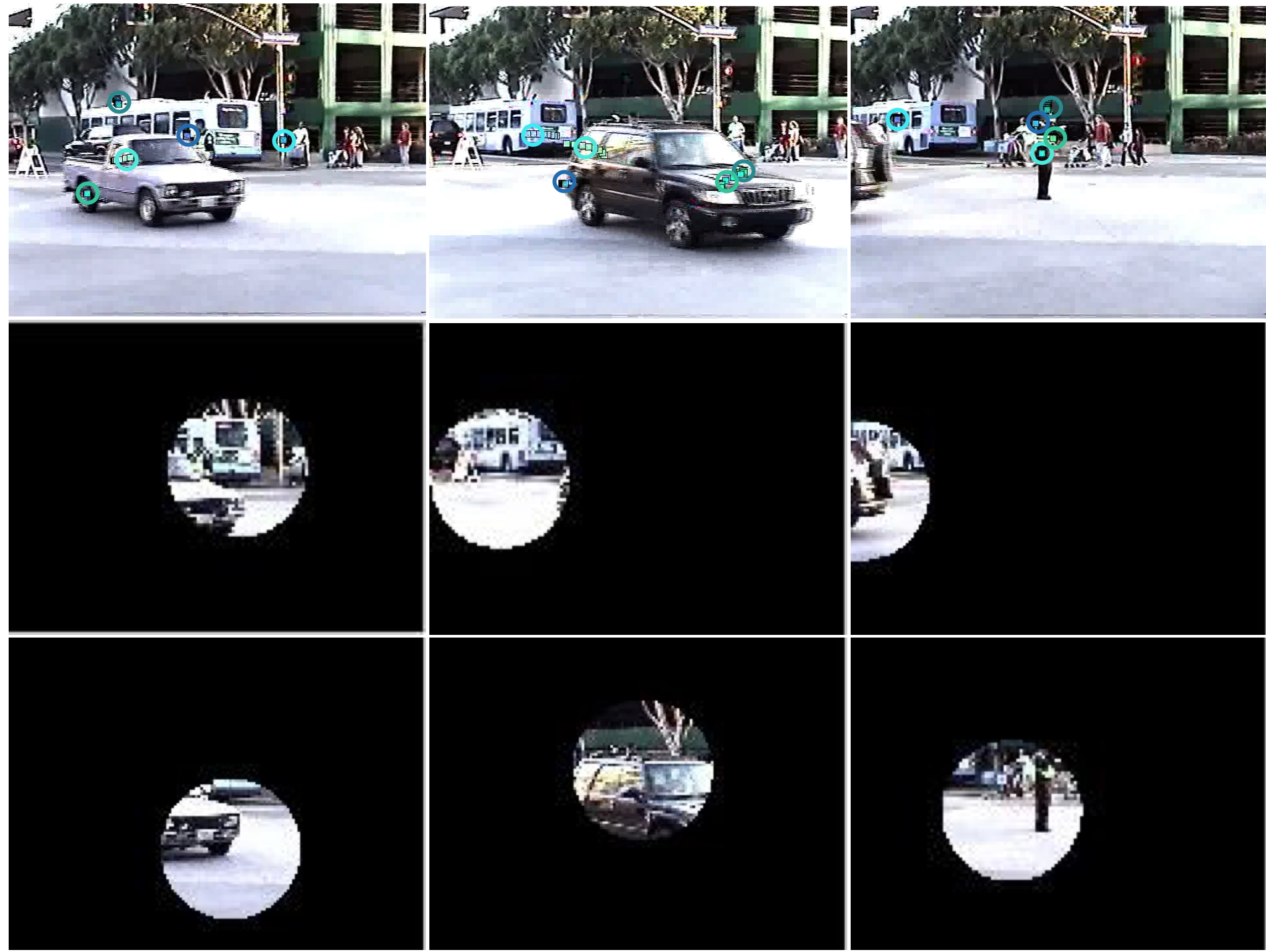}
\caption{Top row: gaze positions (centers of the enhanced colored circles ) recorded from different observers on different frames of the \texttt{monica03} clip  from the CRCNS eye-1 dataset, University of South California (freely available online). Middle and bottom rows show, on the same frames, the fovea position of two ``observers'' simulated by the method described in \cite{BocFerSMCB2013}.}
\label{Fig:monica}
\end{figure}

The sampling scheme proposed is a general one and can be instantiated in different ways. 
For instance, in \cite{BocFerSMCB2013}   the sampling step of Eq. \ref{eq:step3} is performed through a generalization of the Langevin SDE equation used in \cite{BocFerAnnals2012}.   Biases and variability are accounted for by the stochastic component of the equation. Since dealing with image sequences, the SDE provides operates in different dynamic modes: pursuit needs to be taken also into account in addition to saccades and fixational movements. Each mode is governed by a specific set of  parameters of the $\alpha$-stable distribution  estimated from eye tracking data. The choice among modes is accomplished by generalizing the method proposed in \cite{BocFerSMCB2013}, using a Multinoulli distribution on $z(t)$ and with parameters $\pi(t)$ sampled from the conjugate prior, the Dirichlet distribution. In addition, the sampling step is used to  accomplish an internal simulation step, where a number of candidates shifts is proposed and the most convenient is selected according to a decision rule. 
Also,  the gaze dependent perception $\mathcal{W}$ can be modeled at any level of complexity (cfr. example in Figure \ref{Fig:levels}, Section \ref{sec:emo}). In \citeN{BocFerSMCB2013} it is based on proto-objects sampled from time-varying saliency\footnote{Matlab simulation is available for download at \url{https://www.researchgate.net/publication/290816849_Ecological_sampling_of_gaze_shifts_Matlab_code}}. Figure \ref{Fig:monica}
shows an excerpt of  typical results of the model simulation, which compares human variability in gazing (top row) with that of two ``simulated observers'' (bottom rows) while viewing  the \texttt{monica03} clip  from the CRCNS eye-1 dataset.

In \citeN{BocCOGN2014} and \citeN{napboc_TIP2015} the perceptual evaluation component is extended to handle objects and task (external goal) levels by expanding on \citeN{Poggio2010} (cfr. Figure \ref{Fig:models}), and the decision rule $\mathcal{D}(t)$ concerning the selection of the gaze is based on the expected reward according to the given goal $\mathcal{G}$ (see Section \ref{sec:emo}).

Meanwhile, nothing prevents to conceive  more general perceptual evaluation and executive  control components, by considering perceptual and action modalities other than the visual ones in the vein of  \citeN{coen2009visuomotor,cagli2008draughtsman}, where eye movements and hand actions have been coupled with the goal of performing a drawing task.

But most important, the action-perception cycle is by and large conceived in the foraging framework (see \citeNP{bartumeus2009optimal}, for a thorough introduction), which at the most general level is  summarized  in  Table \ref{metaphor}. Visual foraging corresponds to the  time-varying overt deployment of  visual attention achieved through oculomotor actions, namely, gaze shifts. The forager  feeds on patchily distributed preys or resources, spends its  time traveling between patches or searching and handling food within patches.
 While  searching, it gradually depletes the food, hence, the benefit of staying in the patch is likely to gradually diminish with time.  Moment to moment, striving to maximize its foraging efficiency and energy intake, the forager should make decisions: Which is the best patch  to search?    Which prey, if any, should be chased within the patch? When to leave the current  patch for a richer one?

\begin{figure}[t]
\centering
\includegraphics[scale=0.5,keepaspectratio=true]{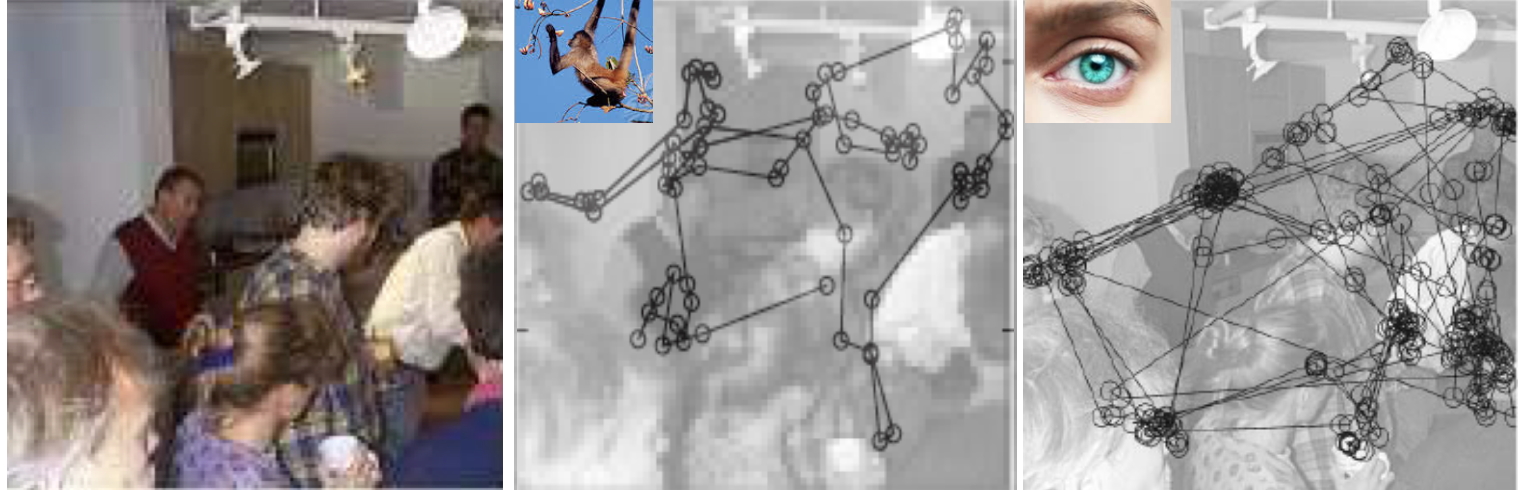}
\caption{Monkey or human: can you tell the difference? The center image has been obtained by superimposing a typical trajectory of spider monkeys foraging in the forest of the Mexican Yucatan, as derived from \cite{ramos2004levy}, on the ``party picture'' (left image) used in \cite{brockgeis}. The right image is an actual human scan path (modified after \citeNP{brockgeis}) }
\label{Fig:spider}
\end{figure}

The spatial behavioral patterns exhibited by foraging animals (but also those detected in human mobility data) are remarkably close to those generated by  gaze shifts \cite{viswanathan2011physics}. Figure \ref{Fig:spider}  presents an intriguing example in this respect.

\begin{figure}[!t]
\centering
\includegraphics[scale= 0.4]{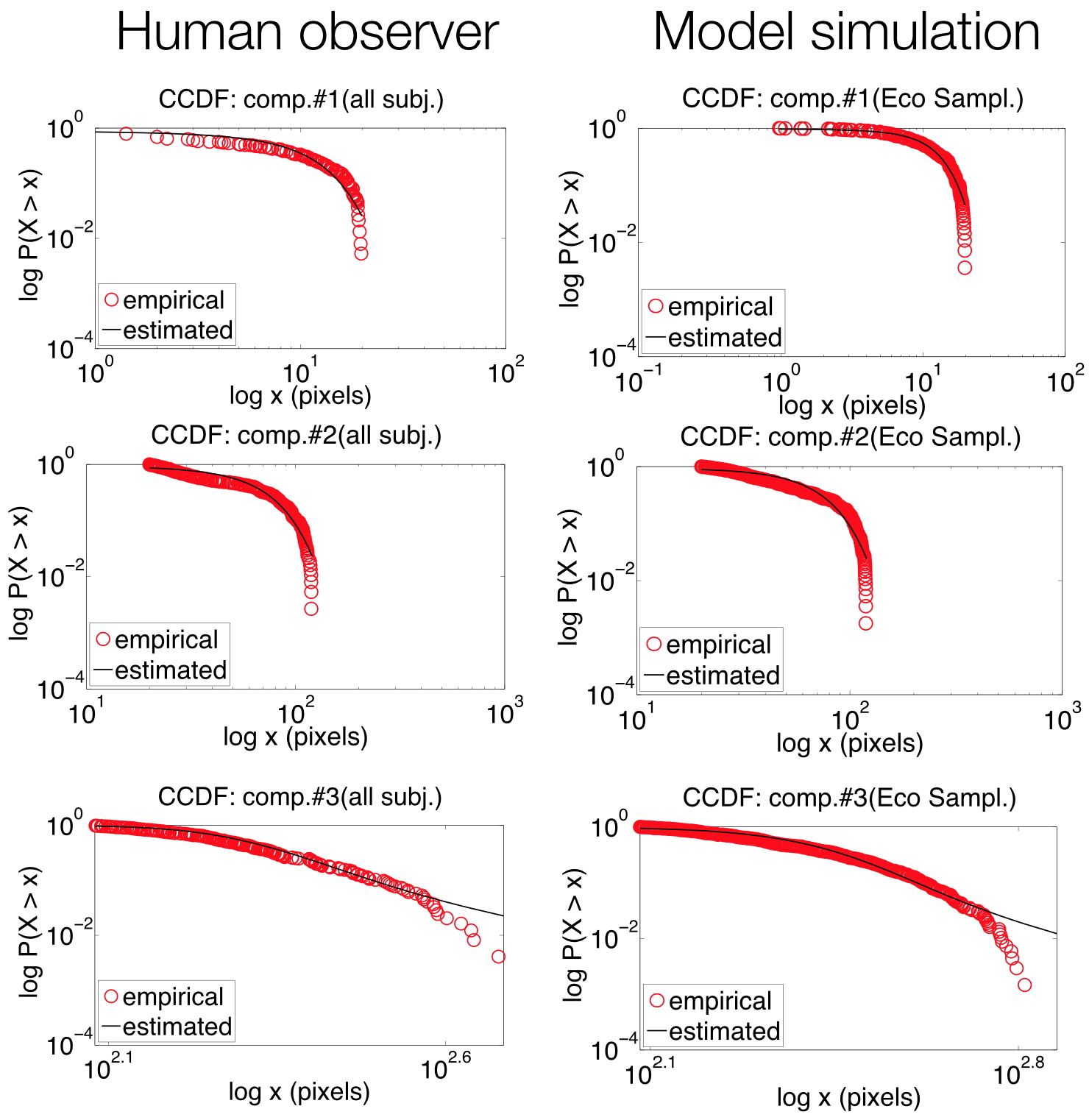}
\caption{Analysis of gaze shift dynamics from the \textsf{monica03} video. The left column shows the double log plot of the CCDF obtained from the gaze magnitude distribution of all subjects: i) the smooth-pursuit mode (top row); ii) the small saccade mode (middle); iii) large saccades (bottom). The right column  presents the equivalent  CCDFs obtained via model simulation. }
\label{Fig:ccdf}
\end{figure}

The fact that the  physics underlying foraging overlaps with that of several other kinds of complex random searches and stochastic optimization problems \cite{viswanathan2011physics},  and notably with that of visual exploration via gaze shifts \cite{viswanathan2011physics,marlow2015temporal,brockgeis},  makes available a variety of analytical  tools beyond classic metrics exploited in computer vision or psychology. For instance, in Figure \ref{Fig:ccdf}, it is shown how the gaze shift amplitude modes from human observers can be compared with those generated via simulation by using the complementary Cumulative Distribution Function (CCDF),   which provides a precise description of  the
distribution of  the gaze shift by considering its upper tail behavior. This  can be defined   as $\overline{F}(x)=P(| \mathbf{r}|>x)=1 - F(x)$, where $F$ is the cumulative distribution function (CDF) of amplitudes.  Consideration of the upper
tail, i.e. the CCDF of jump lengths is a  standard convention in foraging, human mobility,  and anomalous diffusion research \cite{viswanathan2011physics}.

To sum up, foraging offers a novel perspective for formulating models and related evaluations of visual attention and oculomotor behavior \cite{wolfe2013time}. Unifying hypotheses such as the oculomotor continuum from exploration to fixation by \citeN{otero2013oculomotor} can be reconsidered in the light  of fundamental theorems of statistical mechanics \cite{weron2010generalization}.

\begin{table}[t]
\tbl{Relationship between Attentive vision and Foraging}{
\begin{tabular}{|c|c|}
\hline
\textbf{Video stream attentive processing} & \textbf{Patchy landscape foraging} \\
\hline
Observer & Forager \\
\hline
Observer's gaze shift & Forager's relocation \\
\hline
Region of interest  & Patch\\
\hline
Proto-object & Candidate prey\\
\hline
Detected object & Prey\\
\hline
Region selection & Patch choice \\
\hline
Deploying attention to object  & Prey choice and handling\\
\hline
Disengaging from object  & Prey leave\\
\hline
Region leave & Patch leave or giving-up \\
\hline
\end{tabular}}
\label{metaphor}
\end{table}%

Interestingly enough, the reformulation of visual attention in terms of foraging theory is not simply an  informing metaphor.  It has been argued that what was once foraging for tangible resources in a physical space  became, over evolutionary time, foraging in cognitive space for information related to those resources \cite{hills2006animal}, and such adaptations play a fundamental role in goal-directed deployment of visual attention \cite{wolfe2013time}.

\section{Bringing value into the game: a doorway to affective modulation}
\label{sec:emo}
The introduction of a goal level, either exogenous (originating from outside the observer's organism) or endogenous (internal) is not an innocent shift.

From a classical cognitive perspective,  the assignment of  a task to the observer implicitly  defines a value for every point of the space, in the sense that information in some points is more relevant than in others  for the completion of the task;  the shifting  of the gaze on a particular point, in turn,
determines the payoff that can be gained. 

There is a number of psychological and neurobiological studies  showing the availability of value maps and loci of reward influencing the final gaze shift \cite{platt1999neural,leon1999effect,ikeda2003reward,Hikosaka2006}.
The payoff  is nothing else that the value, with respect to the completion of the task,  obtained by moving the fovea in a given position. Thus points associated with high values produce, when fixated,  high payoffs since these fixations bring the observer closer to her/his goal. For  instance, in \cite{BocCOGN2014} and in \cite{napboc_TIP2015}, reward was introduced to make a choice among the candidate gaze shifts stochastically sampled according to Eq.~\ref{eq:step3},  in terms of expected reward (e.g., tuned by the probability of finding the task-assigned  object). 

\begin{figure}[t]
\centering
\includegraphics[scale=0.40,keepaspectratio=true]{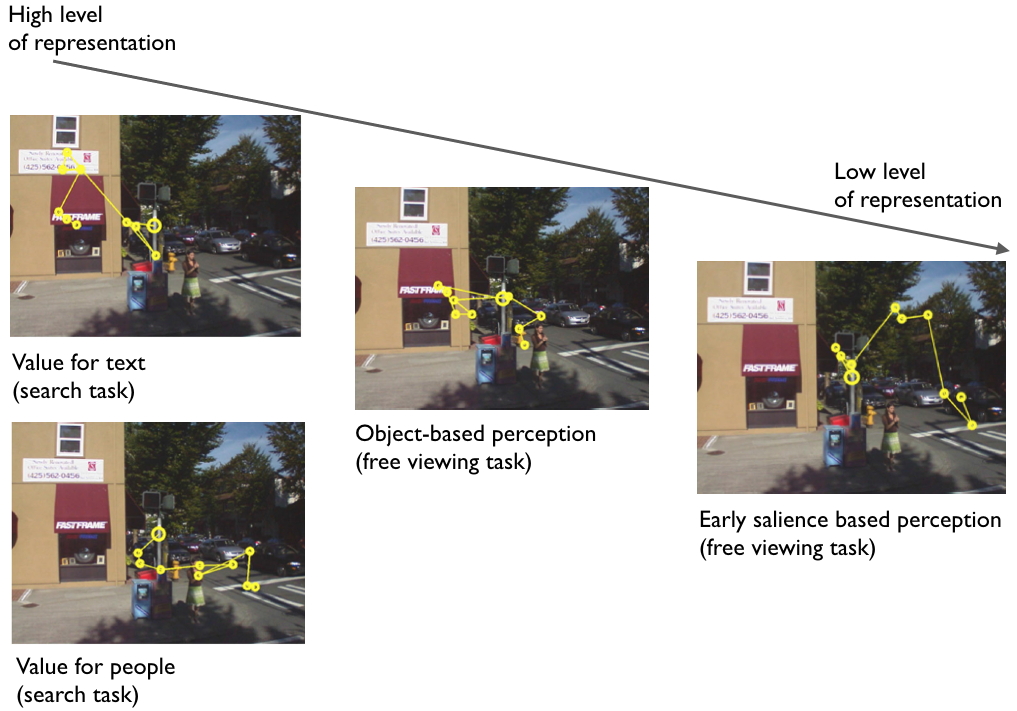}
\caption{Different scan paths originated from the progressive reduction of   representation levels in the perceptual evaluation component presented in Figure \ref{fig:myPGM}.}
\label{Fig:levels}
\end{figure}

Figure \ref{Fig:levels} presents  one example of the different scan paths obtained by progressively reducing  the levels of representation in the perceptual evaluation component presented in Figure \ref{fig:myPGM} \cite{BocCOGN2014,napboc_TIP2015}.  

Value set by the given goal can weight differently the objects within the scene, thus purposively biasing the scan path. Gaze is still uniformly deployed to relevant items within the scene (people, text) when object-based representation is exploited.  When salience alone is used, the generated scan path fails in accounting for the relevant items and bears no relation with the semantics which can be attributed to the scene.

The use of value and reward endow attentive models with the capability of handling complex task. For instance, \citeN{BocCOGN2014} considered the goal of  text spotting in unconstrained urban environments, and its validation encompassed data gathered from a mobile eye tracking device \cite{clav2014}. \citeN{napboc_TIP2015} extended the foraging framework to cope with the difficult problem of attentive monitoring of multiple video streams in a videosurveillance setting.

Yet, developing eye guidance models based on reward is a difficult endeavour and computational models that use reward and uncertainty as central components are still in their  infancy (but see the discussion by \citeNP{TatlerBallard2011eye}). 
In this respect the remarkable work by Ballard and collegues counters the stream. 
Whilst salience, proto-objects and objects are representations  that have been largely addressed in the context of human eye movements, albeit with different emphasis, in contrast,  value  has been neglected  until recently \cite{schutz2011eye}. One reason is that in the real world there is seldom direct payoff (no orange juice for a primary reward) for making good eye movements or punishment for bad ones.

However, the high attentional priority of ecologically pertinent stimuli can also be explained by mechanisms that do not implicate learning value through repeated pairings with reward. For example, a bias to attend to socially relevant stimuli is evident from infancy  
\cite{anderson2013value}.
More generally, the selection of stimuli   by attention has important implications for the survival and wellbeing of an organism, and attentional priority reflects the overall value of such selection (see  \citeNP{anderson2013value} for a  discussion).
Indeed,  engaging attention with potentially harmful and beneficial stimuli  guarantees that the relevant ones  are selected early so to gauge the exact nature of the potential threat or opportunity and to readily initiate defensive or approach behavior. 

Under these circumstances,    \citeN{maunsell2004neuronal} has proposed a broad definition of reward, which  includes
``not only the immediate primary rewards, but also other factors: the preference for a novel location or stimulus,  the satisfaction of performing well or the desire to complete a given task." 
Such  definition is consistent with the different psychological facets of reward \cite{berridge2003parsing}:  i) learning (including explicit and implicit knowledge produced by associative conditioning and cognitive processes);  ii) affect or emotion (implicit ``liking'' and conscious pleasure); iii) motivation (implicit incentive salience ``wanting'' and cognitive incentive goals). Thus, value representation level is central to both goal-driven affective and cognitive engagement with stimuli in the outside world.


In this broader perspective, the effort to put value and reward into the game shows his inner worth in that, by accounting for  the many aspects of ``biological value" - salience, significance,  unpredictability, affective content - , it  paves the way to a wider dimension of information processing, as most recent results on the affective modulation of the visual  processing stream advocate  \cite{pessoa2008relationship,pessoa2010emotion}, and to the effective exploitation of computational attention models in the emerging domain of social signal processing \cite{vinciarelli2009social}.

A number of important studies in the psychological literature (see, for a discussion,  \citeNP{calvo2006eye,Humphrey_lamb}) have addressed the relationship between overt attention behavior and emotional content of pictures. Many of them investigate specific issues related to individuals such as trait anxiety, social anxiety, spider phobia, and exploit restricted sets of stimuli such as emotional faces or spiders. 
In turn,  the study by \citeN{calvo2006eye} has exploited natural images and normal  subjects,  demonstrating an emotional bias both in attentional orienting and engagement among normal participants and using a wider range of emotional pictures. Results might be summarised as follows:
i) emotionally pleasant and unpleasant pictures capture attention more readily than  neutral pictures;
 ii) the emotional bias can be observed early in initial orienting and subsequent engagement of attention;
 iii) the early stimulus-driven attentional capture by emotional stimuli can be counteracted by goal-driven control in later stages of picture processing.

Peculiarly relevant to our case, \citeN{Humphrey_lamb} have shown  that visual saliency does influence eye movements, but the effect is reliably reduced when an emotional object is present. Pictures containing negative objects were recognized more accurately and recalled in greater detail, and participants fixated more on negative objects than positive or neutral ones. Initial fixations were more likely to be on emotional objects than more visually salient neutral ones. Consistently with \citeN{calvo2006eye}, the overall result suggest that the processing of emotional features occurs at a very early stage of perception.

As a matter of fact, emotional factors are completely neglected in the realm of computational models of attention and gaze shifts. Some efforts have been spent in the field of social robotics, where motivational drives have an indirect influence on attention by influencing the behavioral context, which, in turn, is used  to directly manipulate the gains of the attention system (e.g. by tuning the gains of different bottom-up saliency map, \citeNP{breazeal2001active}). Yet, beyond these broadly related attempts, taking into account the specific influence of affect on eye-behaviour is not a central concern within  this field (for a wide review, see \citeNP{ferreira2014attentional}).  Recent works in the image processing and pattern recognition community use eye tracking data for  the inverse problem of the recognition of emotional content of images (e.g.,\citeNP{tavakoli2014emotional,Tavakoli_PONE2015}) or implicit tagging of videos \cite{soleymani2012multimodal}. Thus, they do not address the generative problem of how emotional factors contribute to the generation of gaze shifts in visual tasks.

\begin{figure}[t]
\centering
\includegraphics[scale=0.4,keepaspectratio=true]{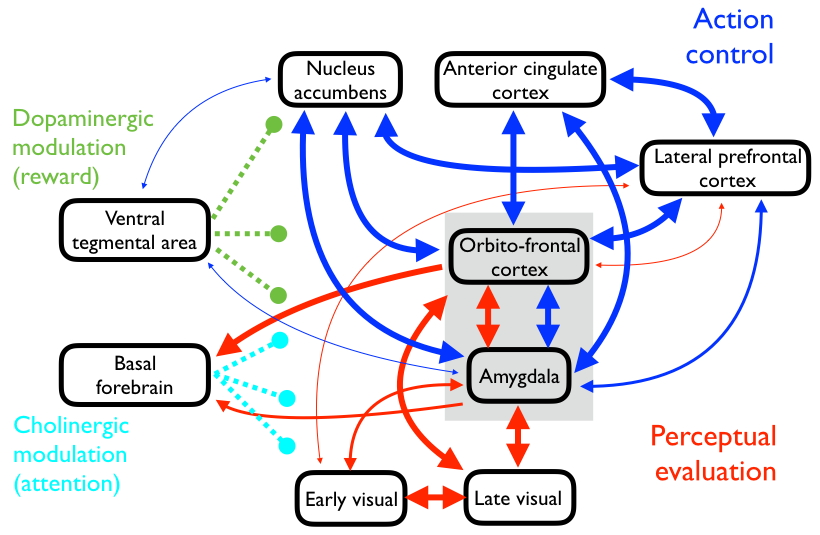}
\caption{Circuits for the processing of visual information and for the executive control \cite{pessoa2008relationship} show that action and perception components are inextricably linked through the mediation of the amygdala and the posterior orbitofrontal cortex (OFC). These  structures are considered  to be the neural substrate of  emotional valence and arousal \cite{salzman2010emotion}. The ventral tegmental area (vTA) and the basal forebrain are responsible for diffuse neuromodulatory effects related to reward and attention. Anterior cingulate cortex (ACC) is likely to be involved in conflict detection and/or error monitoring but also in in computing the benefits and costs of acting  by encoding the probability of reward. Nucleus accumbens is usually related to motivation.The lateral prefrontal cortex (LPFC) plays a central role at the cognitive level in maintaining and manipulating information, but also integrates this content with both affective and motivational information. Line thickness depicts approximate connection strength}
\label{Fig:neuro}
\end{figure}

By contrast, neuroscience has  shown that, crucially, cognitive and emotional contributions cannot be separated, as outlined in Figure \ref{Fig:neuro}.  

The scheme presented, which summarises  an ongoing debate  \cite{pessoa2008relationship}, shows that responses from early and late visual cortex reflecting  stimulus  significance will be a result of simultaneous top-down modulation from fronto-parietal attentional regions (LPFC)  and emotional modulation from the amygdala \cite{Mohanty2013}. On the one hand, stimulus' affective value appears to drive attention and enhance the processing of emotionally modulated information.  On the other hand, exogenously driven attention influences the outcome of affectively significant stimuli \cite{pessoa2008relationship}. As a prominent result, the cognitive or affective origin of the modulation is lost and  stimulus' effect on behaviour is both cognitive and emotional. At the same time, the cognitive control system (LPFC, ACC) guides behaviour while maintaining and manipulating goal-related information; however strategies for action dynamically incorporate value through the mediation of the nucleus accumbens, the amygdala, and the OFC. Eventually,  basal forebrain cholinergic neurons provide regulation of arousal and attention \cite{goard2009basal}, while dopamine neurons located in the vTA  modulate the prediction and expectation of future rewards \cite{pessoa2008relationship}.

\begin{figure}[t]
\centering
\includegraphics[scale=0.4,keepaspectratio=true]{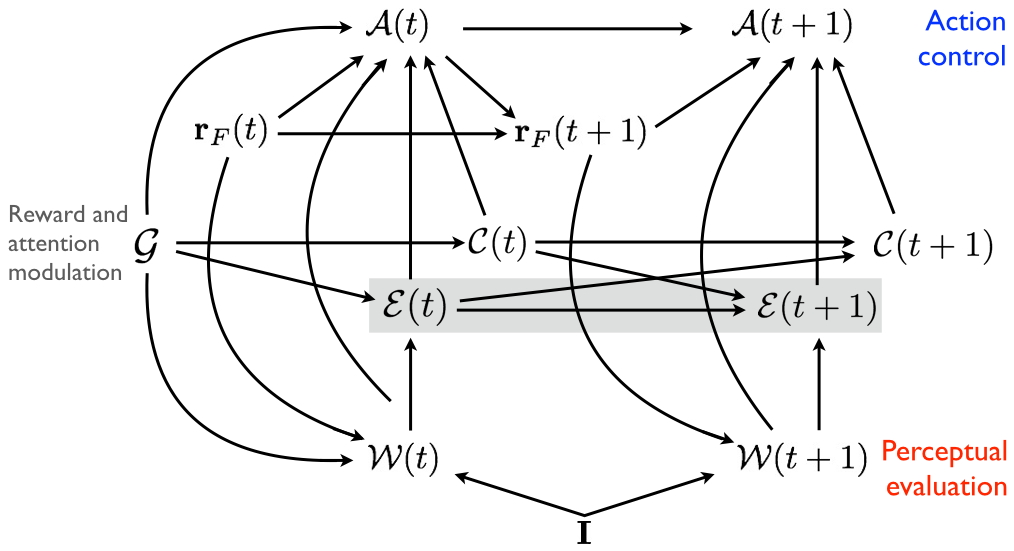}
\caption{The perception-action loop unfolded in time as a dynamic PGM as in Figure \ref{Fig:loop}, but integrated with the core affect latent dimension abstracted as the RV $\mathcal{E}(t)$. $\mathcal{C}(t)$ stands for a  higher cognitive control RV. To simplify the graphics, conditional dependencies $\mathcal{G} \rightarrow \mathcal{A}(t+1)$ and $\mathcal{G} \rightarrow \mathcal{W}(t+1)$ have been omitted.}
\label{Fig:newloop}
\end{figure}

It is to be noted in Figure \ref{Fig:neuro} the central  role of the amygdala and the OFC. It has been argued \cite{salzman2010emotion} that  their tight interaction provides a suitable ground for  representing, at the psychological level the  core affect dimensions \cite{russell2003core} of  valence (pleasure--displeasure conveyed by the visual stimuli) and arousal (activation--deactivation). 
From a computational  standpoint,  the observer's core affect can in principle be modelled as a  dynamic latent space \cite{vitale2014affective}, which we surmise might be readily  embedded within the loop as proposed in Fig. \ref{Fig:newloop}. This way,  gaze shifts   would benefit from the crucial  emotional mediation between the action control and perceptual evaluation components.

\section{Conclusion}
Time is ripe  to abandon the marshland of mass production and evaluation of bottom-up saliency techniques. 
Such boundless effort is partially based on a fatally flawed assumption \cite{santini2008context}:  that  visual data have a meaning \emph{per se}, which can be derived as a function of a certain representation of the data themselves.  Meaning is an outcome of an interpretative process rather than a property of the viewed scene. It is the act of perceiving, contextual and situated, that gives a scene its meaning \cite{wittgenstein2010philosophical}.

The course of modelling  can be more fruitfully directed not only  to climb  the  hierarchy of representation levels and  to cope  with overlooked aspects of eye guidance, but to eventually reappraise the observer within his natural setting: an active observer compelled to  purposively exploit visual attention for accomplishing real-world tasks  \cite{maurice1945phenomenologie}.

\bibliographystyle{ACM-Reference-Format-Journals} 
\bibliography{levyeye}


\begin{thebibliography}{00}


\ifx \showCODEN    \undefined \def \showCODEN     #1{\unskip}     \fi
\ifx \showDOI      \undefined \def \showDOI       #1{{\tt DOI:}\penalty0{#1}\ }
  \fi
\ifx \showISBNx    \undefined \def \showISBNx     #1{\unskip}     \fi
\ifx \showISBNxiii \undefined \def \showISBNxiii  #1{\unskip}     \fi
\ifx \showISSN     \undefined \def \showISSN      #1{\unskip}     \fi
\ifx \showLCCN     \undefined \def \showLCCN      #1{\unskip}     \fi
\ifx \shownote     \undefined \def \shownote      #1{#1}          \fi
\ifx \showarticletitle \undefined \def \showarticletitle #1{#1}   \fi
\ifx \showURL      \undefined \def \showURL       #1{#1}          \fi

\bibitem[\protect\citeauthoryear{Aloimonos, Weiss, and Bandyopadhyay}{Aloimonos
  et~al\mbox{.}}{1988}]%
        {aloimonos1988active}
{John Aloimonos}, {Isaac Weiss}, {and} {Amit Bandyopadhyay}. 1988.
\newblock \showarticletitle{Active vision}.
\newblock {\em International journal of computer vision\/} {1}, 4 (1988),
  333--356.
\newblock


\bibitem[\protect\citeauthoryear{Anderson}{Anderson}{2013}]%
        {anderson2013value}
{Brian~A Anderson}. 2013.
\newblock \showarticletitle{A value-driven mechanism of attentional selection}.
\newblock {\em Journal of vision\/} {13}, 3 (2013).
\newblock


\bibitem[\protect\citeauthoryear{Bajcsy and Campos}{Bajcsy and Campos}{1992}]%
        {bajcsy1992active}
{Ruzena Bajcsy} {and} {Mario Campos}. 1992.
\newblock \showarticletitle{Active and exploratory perception}.
\newblock {\em CVGIP: Image Understanding\/} {56}, 1 (1992), 31--40.
\newblock


\bibitem[\protect\citeauthoryear{Baldi and Itti}{Baldi and Itti}{2010}]%
        {balditti2010}
{P. Baldi} {and} {L. Itti}. 2010.
\newblock \showarticletitle{Of bits and wows: A Bayesian theory of surprise
  with applications to attention}.
\newblock {\em Neural Networks\/} {23}, 5 (2010), 649--666.
\newblock


\bibitem[\protect\citeauthoryear{Ballard}{Ballard}{1991}]%
        {Ballard}
{D.H. Ballard}. 1991.
\newblock \showarticletitle{Animate vision}.
\newblock {\em Artificial intelligence\/} {48}, 1 (1991), 57--86.
\newblock
\showISSN{0004-3702}


\bibitem[\protect\citeauthoryear{Bartumeus and Catalan}{Bartumeus and
  Catalan}{2009}]%
        {bartumeus2009optimal}
{F. Bartumeus} {and} {J. Catalan}. 2009.
\newblock \showarticletitle{{Optimal search behavior and classic foraging
  theory}}.
\newblock {\em Journal of Physics A: Mathematical and Theoretical\/}  {42}
  (2009), 434002.
\newblock


\bibitem[\protect\citeauthoryear{Begum, Karray, Mann, and Gosine}{Begum
  et~al\mbox{.}}{2010}]%
        {begum2010probabilistic}
{M. Begum}, {F. Karray}, {G.K.I. Mann}, {and} {R.G. Gosine}. 2010.
\newblock \showarticletitle{A probabilistic model of overt visual attention for
  cognitive robots}.
\newblock {\em Systems, Man, and Cybernetics, Part B: Cybernetics, IEEE
  Transactions on\/} {40}, 5 (2010), 1305--1318.
\newblock
\showISSN{1083-4419}


\bibitem[\protect\citeauthoryear{Berridge and Robinson}{Berridge and
  Robinson}{2003}]%
        {berridge2003parsing}
{Kent~C Berridge} {and} {Terry~E Robinson}. 2003.
\newblock \showarticletitle{Parsing reward}.
\newblock {\em Trends in neurosciences\/} {26}, 9 (2003), 507--513.
\newblock


\bibitem[\protect\citeauthoryear{Boccignone}{Boccignone}{2008}]%
        {BoccICPR08}
{G. Boccignone}. 2008.
\newblock \showarticletitle{Nonparametric Bayesian attentive video analysis}.
  In {\em Proc. 19th International Conference on Pattern Recognition, ICPR
  2008}. IEEE Press, 1--4.
\newblock


\bibitem[\protect\citeauthoryear{Boccignone and Ferraro}{Boccignone and
  Ferraro}{2004}]%
        {bfpha04}
{G. Boccignone} {and} {M. Ferraro}. 2004.
\newblock \showarticletitle{Modelling gaze shift as a constrained random walk}.
\newblock {\em Physica A: Statistical Mechanics and its Applications\/} {331},
  1-2 (2004), 207--218.
\newblock


\bibitem[\protect\citeauthoryear{Boccignone and Ferraro}{Boccignone and
  Ferraro}{2013}]%
        {BocFerAnnals2012}
{Giuseppe Boccignone} {and} {Mario Ferraro}. 2013.
\newblock \showarticletitle{Feed and fly control of visual scanpaths for
  foveation image processing}.
\newblock {\em annals of telecommunications-annales des
  t{\'e}l{\'e}communications\/} {68}, 3-4 (2013), 201--217.
\newblock


\bibitem[\protect\citeauthoryear{Boccignone and Ferraro}{Boccignone and
  Ferraro}{2014}]%
        {BocFerSMCB2013}
{Giuseppe Boccignone} {and} {Mario Ferraro}. 2014.
\newblock \showarticletitle{Ecological Sampling of Gaze Shifts}.
\newblock {\em {IEEE} Trans. on Cybernetics\/} {44}, 2 (Feb 2014), 266--279.
\newblock


\bibitem[\protect\citeauthoryear{Boccignone, Marcelli, Napoletano, Di~Fiore,
  Iacovoni, and Morsa}{Boccignone et~al\mbox{.}}{2008}]%
        {bocc08tcsvt}
{G. Boccignone}, {A. Marcelli}, {P. Napoletano}, {G. Di~Fiore}, {G. Iacovoni},
  {and} {S. Morsa}. 2008.
\newblock \showarticletitle{{Bayesian integration of face and low-level cues
  for foveated video coding}}.
\newblock {\em IEEE Transactions on Circuits and Systems for Video
  Technology\/} {18}, 12 (2008), 1727--1740.
\newblock


\bibitem[\protect\citeauthoryear{Borji, Cheng, Jiang, and Li}{Borji
  et~al\mbox{.}}{2014}]%
        {borji2014salient}
{Ali Borji}, {Ming-Ming Cheng}, {Huaizu Jiang}, {and} {Jia Li}. 2014.
\newblock \showarticletitle{Salient object detection: A survey}.
\newblock {\em arXiv preprint arXiv:1411.5878\/} (2014).
\newblock


\bibitem[\protect\citeauthoryear{Borji and Itti}{Borji and Itti}{2013}]%
        {BorItti2012}
{A. Borji} {and} {L. Itti}. 2013.
\newblock \showarticletitle{State-of-the-Art in Visual Attention Modeling}.
\newblock {\em IEEE Transactions on Pattern Analysis and Machine
  Intelligence\/} {35}, 1 (2013), 185--207.
\newblock


\bibitem[\protect\citeauthoryear{Borji, Sihite, and Itti}{Borji
  et~al\mbox{.}}{2012}]%
        {borji2012object}
{Ali Borji}, {Dicky~N Sihite}, {and} {Laurent Itti}. 2012.
\newblock \showarticletitle{An Object-Based Bayesian Framework for Top-Down
  Visual Attention}. In {\em Twenty-Sixth AAAI Conference on Artificial
  Intelligence}.
\newblock


\bibitem[\protect\citeauthoryear{Breazeal, Edsinger, Fitzpatrick, and
  Scassellati}{Breazeal et~al\mbox{.}}{2001}]%
        {breazeal2001active}
{Cynthia Breazeal}, {Aaron Edsinger}, {Paul Fitzpatrick}, {and} {Brian
  Scassellati}. 2001.
\newblock \showarticletitle{Active vision for sociable robots}.
\newblock {\em IEEE Transactions on Systems, Man and Cybernetics, Part A:
  Systems and Humans\/} {31}, 5 (2001), 443--453.
\newblock


\bibitem[\protect\citeauthoryear{Brockmann and Geisel}{Brockmann and
  Geisel}{2000}]%
        {brockgeis}
{D. Brockmann} {and} {T. Geisel}. 2000.
\newblock \showarticletitle{{The ecology of gaze shifts}}.
\newblock {\em Neurocomputing\/} {32}, 1 (2000), 643--650.
\newblock


\bibitem[\protect\citeauthoryear{Bruce and Tsotsos}{Bruce and Tsotsos}{2009}]%
        {bruce2009saliency}
{Neil~DB Bruce} {and} {John~K Tsotsos}. 2009.
\newblock \showarticletitle{Saliency, attention, and visual search: An
  information theoretic approach}.
\newblock {\em Journal of vision\/} {9}, 3 (2009), 5--5.
\newblock


\bibitem[\protect\citeauthoryear{Bruce, Wloka, Frosst, Rahman, and
  Tsotsos}{Bruce et~al\mbox{.}}{2015}]%
        {bruce2015computational}
{Neil~DB Bruce}, {Calden Wloka}, {Nick Frosst}, {Shafin Rahman}, {and} {John~K
  Tsotsos}. 2015.
\newblock \showarticletitle{On computational modeling of visual saliency:
  Examining what's right, and what's left}.
\newblock {\em Vision research\/}  {116} (2015), 95--112.
\newblock


\bibitem[\protect\citeauthoryear{Bylinskii, DeGennaro, Rajalingham, Ruda,
  Zhang, and Tsotsos}{Bylinskii et~al\mbox{.}}{2015}]%
        {bylinskii2015towards}
{Z Bylinskii}, {EM DeGennaro}, {R Rajalingham}, {H Ruda}, {J Zhang}, {and} {JK
  Tsotsos}. 2015.
\newblock \showarticletitle{Towards the quantitative evaluation of visual
  attention models}.
\newblock {\em Vision research\/}  {116} (2015), 258--268.
\newblock


\bibitem[\protect\citeauthoryear{Cain, Vul, Clark, and Mitroff}{Cain
  et~al\mbox{.}}{2012}]%
        {cain2012bayesian}
{Matthew~S Cain}, {Edward Vul}, {Kait Clark}, {and} {Stephen~R Mitroff}. 2012.
\newblock \showarticletitle{A Bayesian optimal foraging model of human visual
  search}.
\newblock {\em Psychological science\/} {23}, 9 (2012), 1047--1054.
\newblock


\bibitem[\protect\citeauthoryear{Canosa}{Canosa}{2009}]%
        {canosa2009real}
{R.L. Canosa}. 2009.
\newblock \showarticletitle{Real-world vision: Selective perception and task}.
\newblock {\em ACM Transactions on Applied Perception\/} {6}, 2 (2009), 11.
\newblock


\bibitem[\protect\citeauthoryear{Cerf, Frady, and Koch}{Cerf
  et~al\mbox{.}}{2009}]%
        {cerf2009faces}
{M. Cerf}, {E.P. Frady}, {and} {C. Koch}. 2009.
\newblock \showarticletitle{Faces and text attract gaze independent of the
  task: Experimental data and computer model}.
\newblock {\em Journal of Vision\/} {9}, 12 (2009).
\newblock


\bibitem[\protect\citeauthoryear{Cerf, Harel, Einh{\"a}user, and Koch}{Cerf
  et~al\mbox{.}}{2008}]%
        {cerf2008predicting}
{M. Cerf}, {J. Harel}, {W. Einh{\"a}user}, {and} {C. Koch}. 2008.
\newblock \showarticletitle{Predicting human gaze using low-level saliency
  combined with face detection}.
\newblock {\em Advances in neural information processing systems\/}  {20}
  (2008).
\newblock


\bibitem[\protect\citeauthoryear{Chernyak and Stark}{Chernyak and
  Stark}{2001}]%
        {ChernyakStark}
{D.~A. Chernyak} {and} {L.~W. Stark}. 2001.
\newblock \showarticletitle{Top--Down Guided Eye Movements}.
\newblock {\em {IEEE} Trans. Systems Man Cybernetics - B\/}  {31} (2001),
  514--522.
\newblock


\bibitem[\protect\citeauthoryear{Chikkerur, Serre, Tan, and Poggio}{Chikkerur
  et~al\mbox{.}}{2010}]%
        {Poggio2010}
{S. Chikkerur}, {T. Serre}, {C. Tan}, {and} {T. Poggio}. 2010.
\newblock \showarticletitle{What and where: A Bayesian inference theory of
  attention}.
\newblock {\em Vision research\/} {50}, 22 (2010), 2233--2247.
\newblock
\showISSN{0042-6989}


\bibitem[\protect\citeauthoryear{Clavelli}{Clavelli}{2014}]%
        {clav2014}
{Antonio Clavelli}. 2014.
\newblock {\em A computational model of eye guidance, searching for text in
  real scene images}.
\newblock Ph.D. Dissertation. Universitat Aut\'onoma de Barcelona. Departament
  de Ci\'encies de la Computaci\'o.
\newblock
\showURL{%
\url{http://ddd.uab.cat/record/127173}}


\bibitem[\protect\citeauthoryear{Clavelli, Karatzas, Llad\'{o}s, Ferraro, and
  Boccignone}{Clavelli et~al\mbox{.}}{2014}]%
        {BocCOGN2014}
{Antonio Clavelli}, {Dimosthenis Karatzas}, {Josep Llad\'{o}s}, {Mario
  Ferraro}, {and} {Giuseppe Boccignone}. 2014.
\newblock \showarticletitle{Modelling Task-Dependent Eye Guidance to Objects in
  Pictures}.
\newblock {\em Cognitive Computation\/} {6}, 3 (2014), 558--584.
\newblock


\bibitem[\protect\citeauthoryear{{Coen-Cagli}, Coraggio, Napoletano, and
  Boccignone}{{Coen-Cagli} et~al\mbox{.}}{2008}]%
        {cagli2008draughtsman}
{Ruben {Coen-Cagli}}, {Paolo Coraggio}, {Paolo Napoletano}, {and} {Giuseppe
  Boccignone}. 2008.
\newblock \showarticletitle{What the draughtsman's hand tells the draughtsman's
  eye: A sensorimotor account of drawing}.
\newblock {\em International Journal of Pattern Recognition and Artificial
  Intelligence\/} {22}, 05 (2008), 1015--1029.
\newblock


\bibitem[\protect\citeauthoryear{{Coen-Cagli}, Coraggio, Napoletano, Schwartz,
  Ferraro, and Boccignone}{{Coen-Cagli} et~al\mbox{.}}{2009}]%
        {coen2009visuomotor}
{Ruben {Coen-Cagli}}, {Paolo Coraggio}, {Paolo Napoletano}, {Odelia Schwartz},
  {Mario Ferraro}, {and} {Giuseppe Boccignone}. 2009.
\newblock \showarticletitle{Visuomotor characterization of eye movements in a
  drawing task}.
\newblock {\em Vision research\/} {49}, 8 (2009), 810--818.
\newblock


\bibitem[\protect\citeauthoryear{Cordeschi}{Cordeschi}{2002}]%
        {cordeschi2002discovery}
{Roberto Cordeschi}. 2002.
\newblock {\em The discovery of the artificial: Behavior, mind and machines
  before and beyond cybernetics}. Vol.~28.
\newblock Springer Science \& Business Media.
\newblock


\bibitem[\protect\citeauthoryear{deCroon, Postma, and {van den Herik}}{deCroon
  et~al\mbox{.}}{2011}]%
        {postma2011}
{G.C.H.E. deCroon}, {E.O. Postma}, {and} {H.~J. {van den Herik}}. 2011.
\newblock \showarticletitle{{Adaptive Gaze Control for Object Detection}}.
\newblock {\em Cognitive Computation\/}  {3} (2011), 264--278.
\newblock


\bibitem[\protect\citeauthoryear{Dorr, Martinetz, Gegenfurtner, and Barth}{Dorr
  et~al\mbox{.}}{2010}]%
        {dorr2010variability}
{M. Dorr}, {T. Martinetz}, {K.R. Gegenfurtner}, {and} {E. Barth}. 2010.
\newblock \showarticletitle{Variability of eye movements when viewing dynamic
  natural scenes}.
\newblock {\em Journal of Vision\/} {10}, 10 (2010).
\newblock


\bibitem[\protect\citeauthoryear{Einh\"auser, Spain, and Perona}{Einh\"auser
  et~al\mbox{.}}{2008}]%
        {EinhauserSpainPerona2008}
{Wolfgang Einh\"auser}, {Merrielle Spain}, {and} {Pietro Perona}. 2008.
\newblock \showarticletitle{Objects predict fixations better than early
  saliency}.
\newblock {\em Journal of Vision\/} {8}, 14 (2008).
\newblock
\showDOI{%
\url{http://dx.doi.org/10.1167/8.14.18}}


\bibitem[\protect\citeauthoryear{Elazary and Itti}{Elazary and Itti}{2010}]%
        {elazary2010bayesian}
{Lior Elazary} {and} {Laurent Itti}. 2010.
\newblock \showarticletitle{A Bayesian model for efficient visual search and
  recognition}.
\newblock {\em Vision research\/} {50}, 14 (2010), 1338--1352.
\newblock


\bibitem[\protect\citeauthoryear{Ellis and Stark}{Ellis and Stark}{1986}]%
        {ellistark}
{S.R. Ellis} {and} {L. Stark}. 1986.
\newblock \showarticletitle{{Statistical dependency in visual scanning}}.
\newblock {\em Human Factors: The Journal of the Human Factors and Ergonomics
  Society\/} {28}, 4 (1986), 421--438.
\newblock


\bibitem[\protect\citeauthoryear{Feldman and Friston}{Feldman and
  Friston}{2010}]%
        {feldman2010attention}
{Harriet Feldman} {and} {Karl Friston}. 2010.
\newblock \showarticletitle{Attention, uncertainty, and free-energy}.
\newblock {\em Frontiers in human neuroscience\/}  {4} (2010), 215.
\newblock


\bibitem[\protect\citeauthoryear{Feng}{Feng}{2006}]%
        {feng2006eye}
{G. Feng}. 2006.
\newblock \showarticletitle{Eye movements as time-series random variables: A
  stochastic model of eye movement control in reading}.
\newblock {\em Cognitive Systems Research\/} {7}, 1 (2006), 70--95.
\newblock


\bibitem[\protect\citeauthoryear{Ferreira and Dias}{Ferreira and Dias}{2014}]%
        {ferreira2014attentional}
{Joao~Filipe Ferreira} {and} {Joana Dias}. 2014.
\newblock \showarticletitle{Attentional Mechanisms for Socially Interactive
  Robots--A Survey}.
\newblock {\em IEEE Transactions on Autonomous Mental Development\/} {6}, 2
  (2014), 110--125.
\newblock


\bibitem[\protect\citeauthoryear{Foulsham and Underwood}{Foulsham and
  Underwood}{2008}]%
        {foulsham2008}
{Tom Foulsham} {and} {Geoffrey Underwood}. 2008.
\newblock \showarticletitle{What can saliency models predict about eye
  movements? Spatial and sequential aspects of fixations during encoding and
  recognition}.
\newblock {\em Journal of Vision\/} {8}, 2 (2008).
\newblock


\bibitem[\protect\citeauthoryear{Friston}{Friston}{2010}]%
        {friston2010free}
{Karl Friston}. 2010.
\newblock \showarticletitle{The free-energy principle: a unified brain theory?}
\newblock {\em Nature Reviews Neuroscience\/} {11}, 2 (2010), 127--138.
\newblock


\bibitem[\protect\citeauthoryear{Friston, Schwartenbeck, Fitzgerald,
  Moutoussis, Behrens, and Dolan}{Friston et~al\mbox{.}}{2013}]%
        {friston2013anatomy}
{Karl Friston}, {Philipp Schwartenbeck}, {Thomas Fitzgerald}, {Michael
  Moutoussis}, {Tim Behrens}, {and} {Raymond~J Dolan}. 2013.
\newblock \showarticletitle{The anatomy of choice: active inference and
  agency}.
\newblock {\em Frontiers in Human Neuroscience\/}  {7} (2013), 598.
\newblock


\bibitem[\protect\citeauthoryear{Goard and Dan}{Goard and Dan}{2009}]%
        {goard2009basal}
{Michael Goard} {and} {Yang Dan}. 2009.
\newblock \showarticletitle{Basal forebrain activation enhances cortical coding
  of natural scenes}.
\newblock {\em Nature neuroscience\/} {12}, 11 (2009), 1444--1449.
\newblock


\bibitem[\protect\citeauthoryear{Hacisalihzade, Stark, and Allen}{Hacisalihzade
  et~al\mbox{.}}{1992}]%
        {hacisalihzade1992visual}
{S.S. Hacisalihzade}, {L.W. Stark}, {and} {J.S. Allen}. 1992.
\newblock \showarticletitle{Visual perception and sequences of eye movement
  fixations: A stochastic modeling approach}.
\newblock {\em {IEEE} Trans. Syst., Man, Cybern.\/} {22}, 3 (1992), 474--481.
\newblock
\showISSN{0018-9472}


\bibitem[\protect\citeauthoryear{Harel, Koch, and Perona}{Harel
  et~al\mbox{.}}{2007}]%
        {harel2007graph}
{J. Harel}, {C. Koch}, {and} {P. Perona}. 2007.
\newblock \showarticletitle{Graph-based visual saliency}. In {\em Advances in
  neural information processing systems}, Vol.~19. MIT Press, Cambridge, MA,
  545--552.
\newblock


\bibitem[\protect\citeauthoryear{Hikosaka, Nakamura, and Nakahara}{Hikosaka
  et~al\mbox{.}}{2006}]%
        {Hikosaka2006}
{Okihide Hikosaka}, {Kae Nakamura}, {and} {Hiroyuki Nakahara}. 2006.
\newblock \showarticletitle{Basal Ganglia Orient Eyes to Reward}.
\newblock {\em Journal of Neurophysiology\/} {95}, 2 (2006), 567--584.
\newblock


\bibitem[\protect\citeauthoryear{Hills}{Hills}{2006}]%
        {hills2006animal}
{Thomas~T Hills}. 2006.
\newblock \showarticletitle{Animal Foraging and the Evolution of Goal-Directed
  Cognition}.
\newblock {\em Cognitive Science\/} {30}, 1 (2006), 3--41.
\newblock


\bibitem[\protect\citeauthoryear{Humphrey, Underwood, and Lambert}{Humphrey
  et~al\mbox{.}}{2012}]%
        {Humphrey_lamb}
{Katherine Humphrey}, {Geoffrey Underwood}, {and} {Tony Lambert}. 2012.
\newblock \showarticletitle{Salience of the lambs: A test of the saliency map
  hypothesis with pictures of emotive objects}.
\newblock {\em Journal of Vision\/} {12}, 1 (2012), 22.
\newblock
\showDOI{%
\url{http://dx.doi.org/10.1167/12.1.22}}


\bibitem[\protect\citeauthoryear{Ikeda and Hikosaka}{Ikeda and
  Hikosaka}{2003}]%
        {ikeda2003reward}
{Takuro Ikeda} {and} {Okihide Hikosaka}. 2003.
\newblock \showarticletitle{Reward-dependent gain and bias of visual responses
  in primate superior colliculus}.
\newblock {\em Neuron\/} {39}, 4 (2003), 693--700.
\newblock


\bibitem[\protect\citeauthoryear{Itti and Baldi}{Itti and Baldi}{2009}]%
        {itti2009bayesian}
{L. Itti} {and} {P. Baldi}. 2009.
\newblock \showarticletitle{Bayesian surprise attracts human attention}.
\newblock {\em Vision research\/} {49}, 10 (2009), 1295--1306.
\newblock
\showISSN{0042-6989}


\bibitem[\protect\citeauthoryear{Itti, Koch, and Niebur}{Itti
  et~al\mbox{.}}{1998}]%
        {IttiKoch98}
{L. Itti}, {C. Koch}, {and} {E. Niebur}. 1998.
\newblock \showarticletitle{A Model of Saliency-based Visual Attention for
  Rapid Scene Analysis}.
\newblock {\em IEEE Transactions on Pattern Analysis and Machine
  Intelligence\/}  {20} (1998), 1254--1259.
\newblock


\bibitem[\protect\citeauthoryear{Jiang, Xu, Ye, and Wang}{Jiang
  et~al\mbox{.}}{2015}]%
        {jiang2015image}
{Lai Jiang}, {Mai Xu}, {Zhaoting Ye}, {and} {Zulin Wang}. 2015.
\newblock \showarticletitle{Image Saliency Detection With Sparse Representation
  of Learnt Texture Atoms}. In {\em Proceedings of the IEEE International
  Conference on Computer Vision Workshops}. 54--62.
\newblock


\bibitem[\protect\citeauthoryear{Judd, Ehinger, Durand, and Torralba}{Judd
  et~al\mbox{.}}{2009}]%
        {judd2009learning}
{Tilke Judd}, {Krista Ehinger}, {Fr{\'e}do Durand}, {and} {Antonio Torralba}.
  2009.
\newblock \showarticletitle{Learning to predict where humans look}. In {\em
  IEEE 12th International conference on Computer Vision}. IEEE, 2106--2113.
\newblock


\bibitem[\protect\citeauthoryear{Keech and Resca}{Keech and Resca}{2010}]%
        {keech2010eye1}
{TD Keech} {and} {L. Resca}. 2010.
\newblock \showarticletitle{Eye movements in active visual search: A computable
  phenomenological model}.
\newblock {\em Attention, Perception, \& Psychophysics\/} {72}, 2 (2010),
  285--307.
\newblock


\bibitem[\protect\citeauthoryear{Kienzle, Franz, Sch{\"o}lkopf, and
  Wichmann}{Kienzle et~al\mbox{.}}{2009}]%
        {kienzle2009center}
{Wolf Kienzle}, {Matthias~O Franz}, {Bernhard Sch{\"o}lkopf}, {and} {Felix~A
  Wichmann}. 2009.
\newblock \showarticletitle{Center-surround patterns emerge as optimal
  predictors for human saccade targets}.
\newblock {\em Journal of vision\/} {9}, 5 (2009), 7--7.
\newblock


\bibitem[\protect\citeauthoryear{Kienzle, Wichmann, Franz, and
  Sch{\"o}lkopf}{Kienzle et~al\mbox{.}}{2006}]%
        {kienzle2006nonparametric}
{Wolf Kienzle}, {Felix~A Wichmann}, {Matthias~O Franz}, {and} {Bernhard
  Sch{\"o}lkopf}. 2006.
\newblock \showarticletitle{A nonparametric approach to bottom-up visual
  saliency}. In {\em Advances in neural information processing systems}.
  689--696.
\newblock


\bibitem[\protect\citeauthoryear{Kimura, Pang, Takeuchi, Yamato, and
  Kashino}{Kimura et~al\mbox{.}}{2008}]%
        {kimura2008dynamic}
{A. Kimura}, {D. Pang}, {T. Takeuchi}, {J. Yamato}, {and} {K. Kashino}. 2008.
\newblock \showarticletitle{Dynamic Markov random fields for stochastic
  modeling of visual attention}. In {\em Proc. ICPR '08.} IEEE, 1--5.
\newblock


\bibitem[\protect\citeauthoryear{Kowler}{Kowler}{2011}]%
        {Kowler2011}
{E. Kowler}. 2011.
\newblock \showarticletitle{Eye movements: The past 25 years}.
\newblock {\em Vision Research\/} {51}, 13 (2011), 1457--1483.
\newblock
\newblock
\shownote{50th Anniversary Special Issue of Vision Research - Volume 2.}


\bibitem[\protect\citeauthoryear{Krizhevsky, Sutskever, and Hinton}{Krizhevsky
  et~al\mbox{.}}{2012}]%
        {AlexNetNIPS2012}
{Alex Krizhevsky}, {Ilya Sutskever}, {and} {Geoffrey~E. Hinton}. 2012.
\newblock \showarticletitle{ImageNet Classification with Deep Convolutional
  Neural Networks}.
\newblock In {\em Advances in Neural Information Processing Systems 25},
  {F.~Pereira}, {C.~J.~C. Burges}, {L.~Bottou}, {and} {K.~Q. Weinberger}
  (Eds.). Curran Associates, Inc., 1097--1105.
\newblock
\showURL{%
\url{http://papers.nips.cc/paper/4824-imagenet-classification-with-deep-convolutional-neural-networks.pdf}}


\bibitem[\protect\citeauthoryear{Kruthiventi, Ayush, and Babu}{Kruthiventi
  et~al\mbox{.}}{2015}]%
        {kruthiventi2015deepfix}
{Srinivas~SS Kruthiventi}, {Kumar Ayush}, {and} {R~Venkatesh Babu}. 2015.
\newblock \showarticletitle{DeepFix: A Fully Convolutional Neural Network for
  predicting Human Eye Fixations}.
\newblock {\em arXiv preprint arXiv:1510.02927\/} (2015).
\newblock


\bibitem[\protect\citeauthoryear{K{\"u}mmerer, Theis, and Bethge}{K{\"u}mmerer
  et~al\mbox{.}}{2014}]%
        {kummerer2014deep}
{Matthias K{\"u}mmerer}, {Lucas Theis}, {and} {Matthias Bethge}. 2014.
\newblock \showarticletitle{Deep Gaze I: Boosting saliency prediction with
  feature maps trained on ImageNet}.
\newblock {\em arXiv preprint arXiv:1411.1045\/} (2014).
\newblock


\bibitem[\protect\citeauthoryear{K{\"u}mmerer, Wallis, and Bethge}{K{\"u}mmerer
  et~al\mbox{.}}{2015}]%
        {kummerer2015information}
{Matthias K{\"u}mmerer}, {Thomas~SA Wallis}, {and} {Matthias Bethge}. 2015.
\newblock \showarticletitle{Information-theoretic model comparison unifies
  saliency metrics}.
\newblock {\em Proceedings of the National Academy of Sciences\/} {112}, 52
  (2015), 16054--16059.
\newblock


\bibitem[\protect\citeauthoryear{Lang, Liu, Yu, and Yan}{Lang
  et~al\mbox{.}}{2012}]%
        {lang2012saliency}
{Congyan Lang}, {Guangcan Liu}, {Jian Yu}, {and} {Shuicheng Yan}. 2012.
\newblock \showarticletitle{Saliency detection by multitask sparsity pursuit}.
\newblock {\em IEEE Transactions on Image Processing\/} {21}, 3 (2012),
  1327--1338.
\newblock


\bibitem[\protect\citeauthoryear{Le~Meur and Coutrot}{Le~Meur and
  Coutrot}{2016}]%
        {le2016introducing}
{Olivier Le~Meur} {and} {Antoine Coutrot}. 2016.
\newblock \showarticletitle{Introducing context-dependent and spatially-variant
  viewing biases in saccadic models}.
\newblock {\em Vision Research\/}  {121} (2016), 72--84.
\newblock


\bibitem[\protect\citeauthoryear{Le~Meur and Liu}{Le~Meur and Liu}{2015}]%
        {le2015saccadic}
{Olivier Le~Meur} {and} {Zhi Liu}. 2015.
\newblock \showarticletitle{Saccadic model of eye movements for free-viewing
  condition}.
\newblock {\em Vision research\/}  {116} (2015), 152--164.
\newblock


\bibitem[\protect\citeauthoryear{LeCun, Bengio, and Hinton}{LeCun
  et~al\mbox{.}}{2015}]%
        {lecun2015deep}
{Yann LeCun}, {Yoshua Bengio}, {and} {Geoffrey Hinton}. 2015.
\newblock \showarticletitle{Deep learning}.
\newblock {\em Nature\/} {521}, 7553 (2015), 436--444.
\newblock


\bibitem[\protect\citeauthoryear{Leon and Shadlen}{Leon and Shadlen}{1999}]%
        {leon1999effect}
{Matthew~I Leon} {and} {Michael~N Shadlen}. 1999.
\newblock \showarticletitle{Effect of expected reward magnitude on the response
  of neurons in the dorsolateral prefrontal cortex of the macaque}.
\newblock {\em Neuron\/} {24}, 2 (1999), 415--425.
\newblock


\bibitem[\protect\citeauthoryear{Lin, Kong, Wang, and Zhuang}{Lin
  et~al\mbox{.}}{2014}]%
        {lin2014saliency}
{Yuetan Lin}, {Shu Kong}, {Donghui Wang}, {and} {Yueting Zhuang}. 2014.
\newblock \showarticletitle{Saliency detection within a deep convolutional
  architecture}. In {\em Workshops at the Twenty-Eighth AAAI Conference on
  Artificial Intelligence}.
\newblock


\bibitem[\protect\citeauthoryear{MacKay}{MacKay}{2002}]%
        {Mackay}
{D.J.C. MacKay}. 2002.
\newblock {\em Information Theory, Inference and Learning Algorithms}.
\newblock Cambridge University Press, Cambridge, UK.
\newblock


\bibitem[\protect\citeauthoryear{Marat, Rahman, Pellerin, Guyader, and
  Houzet}{Marat et~al\mbox{.}}{2013}]%
        {marat2013improving}
{Sophie Marat}, {Anis Rahman}, {Denis Pellerin}, {Nathalie Guyader}, {and}
  {Dominique Houzet}. 2013.
\newblock \showarticletitle{Improving Visual Saliency by Adding ÔFace Feature
  MapÕand ÔCenter BiasÕ}.
\newblock {\em Cognitive Computation\/} {5}, 1 (2013), 63--75.
\newblock


\bibitem[\protect\citeauthoryear{Marlow, Viskontas, Matlin, Boydston, Boxer,
  and Taylor}{Marlow et~al\mbox{.}}{2015}]%
        {marlow2015temporal}
{Colleen~A Marlow}, {Indre~V Viskontas}, {Alisa Matlin}, {Cooper Boydston},
  {Adam Boxer}, {and} {Richard~P Taylor}. 2015.
\newblock \showarticletitle{Temporal Structure of Human Gaze Dynamics Is
  Invariant During Free Viewing}.
\newblock {\em PloS one\/} {10}, 9 (2015), e0139379.
\newblock


\bibitem[\protect\citeauthoryear{Marr}{Marr}{1982}]%
        {Marr}
{D. Marr}. 1982.
\newblock {\em Vision: A Computational Investigation into the Human
  Representation and Processing of Visual Information}.
\newblock W.H. Freeman, New York.
\newblock


\bibitem[\protect\citeauthoryear{Martinez, Lungarella, and Pfeifer}{Martinez
  et~al\mbox{.}}{2008}]%
        {martinezLungarella}
{H. Martinez}, {M. Lungarella}, {and} {R. Pfeifer}. 2008.
\newblock \showarticletitle{{Stochastic Extension to the Attention-Selection
  System for the iCub}}.
\newblock {\em University of Zurich, Tech. Rep\/} (2008).
\newblock


\bibitem[\protect\citeauthoryear{Mathe and Sminchisescu}{Mathe and
  Sminchisescu}{2015}]%
        {mathe2015actions}
{Stefan Mathe} {and} {Cristian Sminchisescu}. 2015.
\newblock \showarticletitle{Actions in the eye: dynamic gaze datasets and
  learnt saliency models for visual recognition}.
\newblock {\em Pattern Analysis and Machine Intelligence, IEEE Transactions
  on\/} {37}, 7 (2015), 1408--1424.
\newblock


\bibitem[\protect\citeauthoryear{Maunsell}{Maunsell}{2004}]%
        {maunsell2004neuronal}
{John~HR Maunsell}. 2004.
\newblock \showarticletitle{Neuronal representations of cognitive state: reward
  or attention?}
\newblock {\em Trends in cognitive sciences\/} {8}, 6 (2004), 261--265.
\newblock


\bibitem[\protect\citeauthoryear{{Merleau--Ponty}}{{Merleau--Ponty}}{1945}]%
        {maurice1945phenomenologie}
{Maurice {Merleau--Ponty}}. 1945.
\newblock {\em Ph{\'e}nom{\'e}nologie de la perception}.
\newblock Gallimard, Paris.
\newblock


\bibitem[\protect\citeauthoryear{Mohanty and Sussman}{Mohanty and
  Sussman}{2013}]%
        {Mohanty2013}
{Aprajita Mohanty} {and} {Tamara~J Sussman}. 2013.
\newblock \showarticletitle{Top-down modulation of attention by emotion}.
\newblock {\em Frontiers in Human Neuroscience\/} {7}, 102 (2013).
\newblock
\showURL{%
\url{http://www.frontiersin.org/human_neuroscience/10.3389/fnhum.2013.00102/full}}


\bibitem[\protect\citeauthoryear{Murphy}{Murphy}{2012}]%
        {murphy2012machine}
{Kevin~P Murphy}. 2012.
\newblock {\em Machine learning: a probabilistic perspective}.
\newblock MIT press, Cambridge, MA.
\newblock


\bibitem[\protect\citeauthoryear{Nagai}{Nagai}{2009a}]%
        {nagai2009bottom}
{Y. Nagai}. 2009a.
\newblock \showarticletitle{{From bottom-up visual attention to robot action
  learning}}. In {\em Proceedings of 8 IEEE International Conference on
  Development and Learning}. IEEE Press.
\newblock


\bibitem[\protect\citeauthoryear{Nagai}{Nagai}{2009b}]%
        {nagai2009stability}
{Y. Nagai}. 2009b.
\newblock \showarticletitle{{Stability and sensitivity of bottom-up visual
  attention for dynamic scene analysis}}. In {\em Proceedings of the 2009
  IEEE/RSJ international conference on Intelligent robots and systems}. IEEE
  Press, 5198--5203.
\newblock


\bibitem[\protect\citeauthoryear{Najemnik and Geisler}{Najemnik and
  Geisler}{2005}]%
        {geisler2005}
{J. Najemnik} {and} {W.S. Geisler}. 2005.
\newblock \showarticletitle{Optimal eye movement strategies in visual search}.
\newblock {\em Nature\/} {434}, 7031 (2005), 387--391.
\newblock
\showISSN{0028-0836}


\bibitem[\protect\citeauthoryear{Napoletano, Boccignone, and Tisato}{Napoletano
  et~al\mbox{.}}{2015}]%
        {napboc_TIP2015}
{P. Napoletano}, {G. Boccignone}, {and} {F. Tisato}. 2015.
\newblock \showarticletitle{Attentive monitoring of multiple video streams
  driven by a Bayesian foraging strategy}.
\newblock {\em {IEEE} Trans. on Image Processing\/} {24}, 11 (Nov. 2015), 3266
  -- 3281.
\newblock


\bibitem[\protect\citeauthoryear{Nummenmaa, Hy{\"o}n{\"a}, and Calvo}{Nummenmaa
  et~al\mbox{.}}{2006}]%
        {calvo2006eye}
{Lauri Nummenmaa}, {Jukka Hy{\"o}n{\"a}}, {and} {Manuel~G Calvo}. 2006.
\newblock \showarticletitle{Eye movement assessment of selective attentional
  capture by emotional pictures.}
\newblock {\em Emotion\/} {6}, 2 (2006), 257.
\newblock


\bibitem[\protect\citeauthoryear{Otero-Millan, Macknik, Langston, and
  Martinez-Conde}{Otero-Millan et~al\mbox{.}}{2013}]%
        {otero2013oculomotor}
{Jorge Otero-Millan}, {Stephen~L Macknik}, {Rachel~E Langston}, {and} {Susana
  Martinez-Conde}. 2013.
\newblock \showarticletitle{An oculomotor continuum from exploration to
  fixation}.
\newblock {\em Proceedings of the National Academy of Sciences\/} {110}, 15
  (2013), 6175--6180.
\newblock


\bibitem[\protect\citeauthoryear{Over, Hooge, Vlaskamp, and Erkelens}{Over
  et~al\mbox{.}}{2007}]%
        {over2007coarse}
{E.A.B. Over}, {I.T.C. Hooge}, {B.N.S. Vlaskamp}, {and} {C.J. Erkelens}. 2007.
\newblock \showarticletitle{Coarse-to-fine eye movement strategy in visual
  search}.
\newblock {\em Vision Research\/}  {47} (2007), 2272--2280.
\newblock


\bibitem[\protect\citeauthoryear{Pessoa}{Pessoa}{2008}]%
        {pessoa2008relationship}
{Luiz Pessoa}. 2008.
\newblock \showarticletitle{On the relationship between emotion and cognition}.
\newblock {\em Nature Reviews Neuroscience\/} {9}, 2 (2008), 148--158.
\newblock


\bibitem[\protect\citeauthoryear{Pessoa and Adolphs}{Pessoa and
  Adolphs}{2010}]%
        {pessoa2010emotion}
{Luiz Pessoa} {and} {Ralph Adolphs}. 2010.
\newblock \showarticletitle{Emotion processing and the amygdala: from a'low
  road'to'many roads' of evaluating biological significance}.
\newblock {\em Nature Reviews Neuroscience\/} {11}, 11 (2010), 773--783.
\newblock


\bibitem[\protect\citeauthoryear{Platt and Glimcher}{Platt and
  Glimcher}{1999}]%
        {platt1999neural}
{Michael~L Platt} {and} {Paul~W Glimcher}. 1999.
\newblock \showarticletitle{Neural correlates of decision variables in parietal
  cortex}.
\newblock {\em Nature\/} {400}, 6741 (1999), 233--238.
\newblock


\bibitem[\protect\citeauthoryear{Privitera}{Privitera}{2006}]%
        {privitera2006scanpath}
{Claudio~M Privitera}. 2006.
\newblock \showarticletitle{The scanpath theory: its definitions and later
  developments}. In {\em Electronic Imaging 2006}. International Society for
  Optics and Photonics, 60570A--60570A.
\newblock


\bibitem[\protect\citeauthoryear{R.-Tavakoli, Atyabi, Rantanen, Laukka,
  Nefti-Meziani, and Heikkil?}{R.-Tavakoli et~al\mbox{.}}{2015}]%
        {Tavakoli_PONE2015}
{Hamed R.-Tavakoli}, {Adham Atyabi}, {Antti Rantanen}, {Seppo~J. Laukka},
  {Samia Nefti-Meziani}, {and} {Janne Heikkil?} 2015.
\newblock \showarticletitle{Predicting the Valence of a Scene from Observers?
  Eye Movements}.
\newblock {\em PLoS ONE\/} {10}, 9 (09 2015), 1--19.
\newblock
\showDOI{%
\url{http://dx.doi.org/10.1371/journal.pone.0138198}}


\bibitem[\protect\citeauthoryear{Ramos-Fernandez, Mateos, Miramontes, Cocho,
  Larralde, and Ayala-Orozco}{Ramos-Fernandez et~al\mbox{.}}{2004}]%
        {ramos2004levy}
{G. Ramos-Fernandez}, {J.L. Mateos}, {O. Miramontes}, {G. Cocho}, {H.
  Larralde}, {and} {B. Ayala-Orozco}. 2004.
\newblock \showarticletitle{{L{\'e}vy walk patterns in the foraging movements
  of spider monkeys (Ateles geoffroyi)}}.
\newblock {\em Behavioral Ecology and Sociobiology\/} {55}, 3 (2004), 223--230.
\newblock


\bibitem[\protect\citeauthoryear{Rao}{Rao}{2005}]%
        {rao2005}
{R.P.N. Rao}. 2005.
\newblock \showarticletitle{Bayesian inference and attentional modulation in
  the visual cortex}.
\newblock {\em Neuroreport\/} {16}, 16 (2005), 1843.
\newblock


\bibitem[\protect\citeauthoryear{Rao, Zelinsky, Hayhoe, and Ballard}{Rao
  et~al\mbox{.}}{2002}]%
        {Rao2002}
{Rajesh~P.N. Rao}, {Gregory~J. Zelinsky}, {Mary~M. Hayhoe}, {and} {Dana~H.
  Ballard}. 2002.
\newblock \showarticletitle{Eye movements in iconic visual search}.
\newblock {\em Vision Research\/} {42}, 11 (2002), 1447 -- 1463.
\newblock


\bibitem[\protect\citeauthoryear{Rensink}{Rensink}{2000}]%
        {rensink2000dynamic}
{R.A. Rensink}. 2000.
\newblock \showarticletitle{The dynamic representation of scenes}.
\newblock {\em Visual Cognition\/} {1}, 3 (2000), 17--42.
\newblock


\bibitem[\protect\citeauthoryear{Rothenstein and Tsotsos}{Rothenstein and
  Tsotsos}{2008}]%
        {rothenstein2008attention}
{Albert~L Rothenstein} {and} {John~K Tsotsos}. 2008.
\newblock \showarticletitle{Attention links sensing to recognition}.
\newblock {\em Image and Vision Computing\/} {26}, 1 (2008), 114--126.
\newblock


\bibitem[\protect\citeauthoryear{Rothkopf, Ballard, and Hayhoe}{Rothkopf
  et~al\mbox{.}}{2007}]%
        {rothkopfBallard2007}
{C.A. Rothkopf}, {D.H. Ballard}, {and} {M.M. Hayhoe}. 2007.
\newblock \showarticletitle{Task and context determine where you look}.
\newblock {\em Journal of Vision\/} {7}, 14 (2007).
\newblock


\bibitem[\protect\citeauthoryear{Russell}{Russell}{2003}]%
        {russell2003core}
{James~A Russell}. 2003.
\newblock \showarticletitle{Core affect and the psychological construction of
  emotion.}
\newblock {\em Psychological review\/} {110}, 1 (2003), 145.
\newblock


\bibitem[\protect\citeauthoryear{Rutishauser and Koch}{Rutishauser and
  Koch}{2007}]%
        {rutishauser2007probabilistic}
{U. Rutishauser} {and} {C. Koch}. 2007.
\newblock \showarticletitle{Probabilistic modeling of eye movement data during
  conjunction search via feature-based attention}.
\newblock {\em Journal of Vision\/} {7}, 6 (2007).
\newblock


\bibitem[\protect\citeauthoryear{Salzman and Fusi}{Salzman and Fusi}{2010}]%
        {salzman2010emotion}
{C~Daniel Salzman} {and} {Stefano Fusi}. 2010.
\newblock \showarticletitle{Emotion, cognition, and mental state representation
  in amygdala and prefrontal cortex}.
\newblock {\em Annual review of neuroscience\/}  {33} (2010), 173.
\newblock


\bibitem[\protect\citeauthoryear{Santini and Dumitrescu}{Santini and
  Dumitrescu}{2008}]%
        {santini2008context}
{Simone Santini} {and} {Alexandra Dumitrescu}. 2008.
\newblock \showarticletitle{Context as a non-ontological determinant of
  semantics}.
\newblock In {\em Semantic Multimedia}. Springer, 121--136.
\newblock


\bibitem[\protect\citeauthoryear{Sch{\"u}tz, Braun, and
  Gegenfurtner}{Sch{\"u}tz et~al\mbox{.}}{2011}]%
        {schutz2011eye}
{A.C. Sch{\"u}tz}, {D.I. Braun}, {and} {K.R. Gegenfurtner}. 2011.
\newblock \showarticletitle{Eye movements and perception: A selective review}.
\newblock {\em Journal of Vision\/} {11}, 5 (2011).
\newblock


\bibitem[\protect\citeauthoryear{Seo and Milanfar}{Seo and Milanfar}{2009}]%
        {seo2009}
{HJ Seo} {and} {P. Milanfar}. 2009.
\newblock \showarticletitle{Static and space-time visual saliency detection by
  self-resemblance}.
\newblock {\em Journal of Vision\/} {9}, 12 (2009), 1--27.
\newblock


\bibitem[\protect\citeauthoryear{Shen and Zhao}{Shen and Zhao}{2014}]%
        {shenZhao2014learning}
{Chengyao Shen} {and} {Qi Zhao}. 2014.
\newblock \showarticletitle{Learning to predict eye fixations for semantic
  contents using multi-layer sparse network}.
\newblock {\em Neurocomputing\/}  {138} (2014), 61--68.
\newblock


\bibitem[\protect\citeauthoryear{Soleymani, Lichtenauer, Pun, and
  Pantic}{Soleymani et~al\mbox{.}}{2012}]%
        {soleymani2012multimodal}
{Mohammad Soleymani}, {Jeroen Lichtenauer}, {Thierry Pun}, {and} {Maja Pantic}.
  2012.
\newblock \showarticletitle{A multimodal database for affect recognition and
  implicit tagging}.
\newblock {\em Affective Computing, IEEE Transactions on\/} {3}, 1 (2012),
  42--55.
\newblock


\bibitem[\protect\citeauthoryear{Srinivas, Sarvadevabhatla, Mopuri, Prabhu,
  Kruthiventi, and Radhakrishnan}{Srinivas et~al\mbox{.}}{2016}]%
        {deepTaxonomy2016}
{Suraj Srinivas}, {Ravi~Kiran Sarvadevabhatla}, {Konda~Reddy Mopuri}, {Nikita
  Prabhu}, {Srinivas Kruthiventi}, {and} {Venkatesh~Babu Radhakrishnan}. 2016.
\newblock \showarticletitle{A taxonomy of Deep Convolutional Neural Nets for
  Computer Vision}.
\newblock {\em Frontiers in Robotics and AI\/} {2}, 36 (2016).
\newblock
\showDOI{%
\url{http://dx.doi.org/10.3389/frobt.2015.00036}}


\bibitem[\protect\citeauthoryear{Strasburger, Rentschler, and
  J{\"u}ttner}{Strasburger et~al\mbox{.}}{2011}]%
        {strasburger2011peripheral}
{Hans Strasburger}, {Ingo Rentschler}, {and} {Martin J{\"u}ttner}. 2011.
\newblock \showarticletitle{Peripheral vision and pattern recognition: A
  review}.
\newblock {\em Journal of Vision\/} {11}, 5 (2011).
\newblock


\bibitem[\protect\citeauthoryear{Sun, Fisher, Wang, and Gomes}{Sun
  et~al\mbox{.}}{2008}]%
        {Sun2008}
{Yaoru Sun}, {Robert Fisher}, {Fang Wang}, {and} {Herman~Martins Gomes}. 2008.
\newblock \showarticletitle{A computer vision model for visual-object-based
  attention and eye movements}.
\newblock {\em Computer Vision and Image Understanding\/} {112}, 2 (2008), 126
  -- 142.
\newblock


\bibitem[\protect\citeauthoryear{Tatler, Hayhoe, Land, and Ballard}{Tatler
  et~al\mbox{.}}{2011}]%
        {TatlerBallard2011eye}
{B.W. Tatler}, {M.M. Hayhoe}, {M.F. Land}, {and} {D.H. Ballard}. 2011.
\newblock \showarticletitle{Eye guidance in natural vision: Reinterpreting
  salience}.
\newblock {\em Journal of vision\/} {11}, 5 (2011).
\newblock


\bibitem[\protect\citeauthoryear{Tatler and Vincent}{Tatler and
  Vincent}{2008}]%
        {tatler2008systematic}
{B.W. Tatler} {and} {B.T. Vincent}. 2008.
\newblock \showarticletitle{Systematic tendencies in scene viewing}.
\newblock {\em Journal of Eye Movement Research\/} {2}, 2 (2008), 1--18.
\newblock


\bibitem[\protect\citeauthoryear{Tatler and Vincent}{Tatler and
  Vincent}{2009}]%
        {tatler2009prominence}
{B.W. Tatler} {and} {B.T. Vincent}. 2009.
\newblock \showarticletitle{The prominence of behavioural biases in eye
  guidance}.
\newblock {\em Visual Cognition\/} {17}, 6-7 (2009), 1029--1054.
\newblock


\bibitem[\protect\citeauthoryear{Tavakoli, Rahtu, and Heikkil{\"a}}{Tavakoli
  et~al\mbox{.}}{2013}]%
        {tavakoli2013stochastic}
{Hamed~Rezazadegan Tavakoli}, {Esa Rahtu}, {and} {Janne Heikkil{\"a}}. 2013.
\newblock \showarticletitle{Stochastic bottom--up fixation prediction and
  saccade generation}.
\newblock {\em Image and Vision Computing\/} {31}, 9 (2013), 686--693.
\newblock


\bibitem[\protect\citeauthoryear{Tavakoli, Yanulevskaya, Rahtu, Heikkila, and
  Sebe}{Tavakoli et~al\mbox{.}}{2014}]%
        {tavakoli2014emotional}
{Hamed~R Tavakoli}, {Victoria Yanulevskaya}, {Esa Rahtu}, {Janne Heikkila},
  {and} {Nicu Sebe}. 2014.
\newblock \showarticletitle{Emotional Valence Recognition, Analysis of Salience
  and Eye Movements}. In {\em 2014 22nd International Conference on Pattern
  Recognition (ICPR)}. IEEE, 4666--4671.
\newblock


\bibitem[\protect\citeauthoryear{Torralba}{Torralba}{2003}]%
        {Torralba}
{A Torralba}. 2003.
\newblock \showarticletitle{Contextual priming for object detection}.
\newblock {\em Int. J. of Comp. Vis.\/}  {53} (2003), 153--167.
\newblock


\bibitem[\protect\citeauthoryear{Torralba, Oliva, Castelhano, and
  Henderson}{Torralba et~al\mbox{.}}{2006}]%
        {torralba2006contextual}
{A. Torralba}, {A. Oliva}, {M.S. Castelhano}, {and} {J.M. Henderson}. 2006.
\newblock \showarticletitle{Contextual guidance of eye movements and attention
  in real-world scenes: the role of global features in object search.}
\newblock {\em Psychological review\/} {113}, 4 (2006), 766.
\newblock


\bibitem[\protect\citeauthoryear{van Beers}{van Beers}{2007}]%
        {vanBeers2007sources}
{R.J. van Beers}. 2007.
\newblock \showarticletitle{The sources of variability in saccadic eye
  movements}.
\newblock {\em The Journal of Neuroscience\/} {27}, 33 (2007), 8757--8770.
\newblock


\bibitem[\protect\citeauthoryear{Vig, Dorr, and Cox}{Vig et~al\mbox{.}}{2014}]%
        {vig2014large}
{Eleonora Vig}, {Michael Dorr}, {and} {David Cox}. 2014.
\newblock \showarticletitle{Large-scale optimization of hierarchical features
  for saliency prediction in natural images}. In {\em Proceedings of the IEEE
  Conference on Computer Vision and Pattern Recognition}. 2798--2805.
\newblock


\bibitem[\protect\citeauthoryear{Vinciarelli, Pantic, and Bourlard}{Vinciarelli
  et~al\mbox{.}}{2009}]%
        {vinciarelli2009social}
{Alessandro Vinciarelli}, {Maja Pantic}, {and} {Herv{\'e} Bourlard}. 2009.
\newblock \showarticletitle{Social signal processing: Survey of an emerging
  domain}.
\newblock {\em Image and Vision Computing\/} {27}, 12 (2009), 1743--1759.
\newblock


\bibitem[\protect\citeauthoryear{Viswanathan, Da~Luz, Raposo, and
  Stanley}{Viswanathan et~al\mbox{.}}{2011}]%
        {viswanathan2011physics}
{Gandhimohan~M Viswanathan}, {Marcos~GE Da~Luz}, {Ernesto~P Raposo}, {and}
  {H~Eugene Stanley}. 2011.
\newblock {\em The physics of foraging: an introduction to random searches and
  biological encounters}.
\newblock Cambridge University Press, Cambridge, UK.
\newblock


\bibitem[\protect\citeauthoryear{Vitale, Williams, Johnston, and
  Boccignone}{Vitale et~al\mbox{.}}{2014}]%
        {vitale2014affective}
{Jonathan Vitale}, {Mary-Anne Williams}, {Benjamin Johnston}, {and} {Giuseppe
  Boccignone}. 2014.
\newblock \showarticletitle{Affective facial expression processing via
  simulation: A probabilistic model}.
\newblock {\em Biologically Inspired Cognitive Architectures Journal\/}  {10}
  (2014), 30--41.
\newblock


\bibitem[\protect\citeauthoryear{Wang, Wang, and Ji}{Wang
  et~al\mbox{.}}{2016}]%
        {wang2016deep}
{Kang Wang}, {Shen Wang}, {and} {Qiang Ji}. 2016.
\newblock \showarticletitle{Deep eye fixation map learning for calibration-free
  eye gaze tracking}. In {\em Proceedings of the Ninth Biennial ACM Symposium
  on Eye Tracking Research \& Applications}. ACM, 47--55.
\newblock


\bibitem[\protect\citeauthoryear{Weron and Magdziarz}{Weron and
  Magdziarz}{2010}]%
        {weron2010generalization}
{Aleksander Weron} {and} {Marcin Magdziarz}. 2010.
\newblock \showarticletitle{Generalization of the {Khinchin} theorem to
  {L{\'e}vy} flights}.
\newblock {\em Physical review letters\/} {105}, 26 (2010), 260603.
\newblock


\bibitem[\protect\citeauthoryear{Wischnewski, Belardinelli, Schneider, and
  Steil}{Wischnewski et~al\mbox{.}}{2010}]%
        {anna}
{Marco Wischnewski}, {Anna Belardinelli}, {Werner Schneider}, {and} {Jochen
  Steil}. 2010.
\newblock \showarticletitle{{Where to Look Next? Combining Static and Dynamic
  Proto-objects in a TVA-based Model of Visual Attention}}.
\newblock {\em Cognitive Computation\/} {2}, 4 (2010), 326--343.
\newblock


\bibitem[\protect\citeauthoryear{Wittgenstein}{Wittgenstein}{2010}]%
        {wittgenstein2010philosophical}
{Ludwig Wittgenstein}. 2010.
\newblock {\em Philosophical investigations}.
\newblock John Wiley \& Sons.
\newblock


\bibitem[\protect\citeauthoryear{Wolfe}{Wolfe}{2013}]%
        {wolfe2013time}
{Jeremy~M Wolfe}. 2013.
\newblock \showarticletitle{When is it time to move to the next raspberry bush?
  {F}oraging rules in human visual search}.
\newblock {\em Journal of vision\/} {13}, 3 (2013), 10.
\newblock


\bibitem[\protect\citeauthoryear{Yan, Zhu, Liu, and Liu}{Yan
  et~al\mbox{.}}{2010}]%
        {yan2010visual}
{Junchi Yan}, {Mengyuan Zhu}, {Huanxi Liu}, {and} {Yuncai Liu}. 2010.
\newblock \showarticletitle{Visual saliency detection via sparsity pursuit}.
\newblock {\em Signal Processing Letters, IEEE\/} {17}, 8 (2010), 739--742.
\newblock


\bibitem[\protect\citeauthoryear{Yu, Zhao, Tian, and Tan}{Yu
  et~al\mbox{.}}{2014}]%
        {yu2014maximal}
{Jin-Gang Yu}, {Ji Zhao}, {Jinwen Tian}, {and} {Yihua Tan}. 2014.
\newblock \showarticletitle{Maximal entropy random walk for region-based visual
  saliency}.
\newblock {\em IEEE Transactions on Cybernetics\/} {44}, 9 (2014), 1661--1672.
\newblock


\bibitem[\protect\citeauthoryear{Zelinsky}{Zelinsky}{2008}]%
        {zelinsky2008theory}
{Gregory~J Zelinsky}. 2008.
\newblock \showarticletitle{A theory of eye movements during target
  acquisition.}
\newblock {\em Psychological review\/} {115}, 4 (2008), 787.
\newblock


\end{thebibliography}

\end{document}